\documentclass{article}

\usepackage{microtype}
\usepackage{graphicx}
\usepackage{subcaption}
\usepackage{booktabs} 
\usepackage{multirow}
\usepackage{tabularx} 
\usepackage{enumitem}
\usepackage{hyperref}

\usepackage[preprint]{icml2026}

\usepackage{amsmath}
\usepackage{amssymb}
\usepackage{mathtools}
\usepackage{amsthm}
\definecolor{canyon}{RGB}{205, 92, 92}

\usepackage[capitalize,noabbrev]{cleveref}

\theoremstyle{plain}

\theoremstyle{definition}

\theoremstyle{remark}

\usepackage[textsize=tiny]{todonotes}

\icmltitlerunning{}

\begin{document}

\twocolumn[

  \icmltitle{RiboBridge: protein-conditioned de novo RNA binder structure design}

  \icmlsetsymbol{equal}{*}

  \begin{icmlauthorlist}
    \icmlauthor{Zhou Zhang}{sch1}
    \icmlauthor{Hanqun Cao}{sch2}
    \icmlauthor{Minghao Sun}{sch3}
    \icmlauthor{Odin Zhang}{sch2}
    \icmlauthor{Yang Zhang}{sch3}
    \icmlauthor{Pheng Ann Heng}{sch2}
    \icmlauthor{Wujun Li}{sch1}
    \icmlauthor{Tianfan Fu}{sch1}
  \end{icmlauthorlist}

  \icmlaffiliation{sch1}{State Key Laboratory for Novel Software Technology at Nanjing University, School of Computer Science, Nanjing University}
  \icmlaffiliation{sch2}{The Chinese University of Hong Kong}
  \icmlaffiliation{sch3}{National University of Singapore}

  \icmlcorrespondingauthor{Hanqun Cao}{hqcao23@cse.cuhk.edu.hk}
  \icmlcorrespondingauthor{Tianfan Fu}{futianfan@nju.edu.cn}

  \icmlkeywords{Machine Learning, ICML}

  \vskip 0.3in
]


\printAffiliationsAndNotice{}  

\begin{abstract}
Designing RNA molecules that bind to target protein structures requires generating both an RNA conformation and its bound complex geometry, without assuming a predefined binding site. Existing approaches generally model these coupled processes within a unified representation, entangling RNA conformation generation with protein-relative placement.
In this work, we introduce \textbf{RiboBridge}, a structure-first framework that separates RNA conformational generation from protein--RNA assembly through cross-frame interaction geometry. RiboBridge first proposes multiple candidate RNA-binding interfaces on the protein surface, generates canonical RNA conformations within a pretrained RNA structural latent space (\textbf{RiboSphere}), and then predicts protein--RNA interaction constraints to bridge the independently defined RNA and protein coordinate frames. The resulting interaction geometry is converted into a geometrically consistent protein-relative pose through
rigid placement.
We evaluate RiboBridge on a temporally split, sequence-filtered, and redundancy-controlled benchmark of 124 protein--RNA complexes. Under post-hoc interface-correspondence evaluation, where native information is used only after generation to select among already generated predicted-pocket candidates, RiboBridge achieves the highest fNAT and DockQ and the lowest interface and ligand RMSDs among compared methods. Component ablations and stage-wise diagnostics further characterize the contributions of RNA conformational generation, interaction prediction, and rigid placement, identifying the main sources of complex-recovery error.
\end{abstract}

\section{Introduction}
\label{sec:introduction}
Designing a protein-binding RNA based on a target protein structure differs fundamentally from both RNA folding~\citep{shen2024accurate} and conventional molecular docking~\citep{yan2017hdock}. In many structure-guided molecular design settings, the binding region of the target protein is provided or can be specified before generating a ligand~\citep{zhang2025odesign, peng2022pocket2mol}. The design problem then focuses on constructing a molecule that adopts a compatible structure within a known interaction environment. Here, we consider a site-agnostic setting in which neither the RNA molecule nor its binding interface on the protein is given. The unknown interface leaves both the contact region and the spatial context for RNA placement unresolved. Given only a target protein structure and a desired RNA length, our goal is to generate candidate RNA structures without access to the native RNA sequence, native RNA structure, or a user-specified binding site.

The difficulty arises because protein–RNA complex generation requires both an RNA conformation and a compatible protein-relative pose. RNA exhibits intrinsic structural regularities~\citep{anand2025rna}, while its bound conformation must also accommodate the target interaction environment. Direct generation in a shared complex frame couples conformational modeling, interface discovery, and relative placement. Given the limited availability of experimentally resolved protein–RNA complexes~\citep{han2025copra,szikszai2024rna3db}, this motivates reusing a pretrained RNA structural prior while learning target-conditioned conformations and their assembly with the protein.

Existing approaches address protein-conditioned RNA design from complementary perspectives. Sequence-first methods generate or optimize RNA sequences and subsequently rely on structure prediction or experimental selection to identify functional candidates~\citep{iwano2022generative,wang2024aptadiff,shukueian2025rnatranslator,roman2025bang}. Structure-aware approaches improve access to RNA geometries by generating RNA backbones, joint sequence--structure representations, or protein--nucleic acid complexes~\citep{nori2024rnaflow,abir2025rnaefm,favor2025novo,zhang2025odesign}. However, these approaches generally model RNA conformation and protein-relative placement within a unified complex representation.

We introduce \textbf{RiboBridge}, a structure-first framework that separates pocket-conditioned RNA conformational generation from protein-relative assembly. Given a target protein and a desired RNA length, it proposes candidate binding interfaces and then generates an RNA conformation for each within the pretrained structural space of \textbf{RiboSphere}~\citep{zhang2026ribosphere}. These conformations retain interface-specific conditioning but have no specified pose relative to the protein. The central challenge is then to infer interaction geometry between structures whose relative rotation and translation are unknown. RiboBridge addresses this by predicting bound-state protein–RNA distance distributions from independently encoded within-chain geometry, without using pre-placement cross-chain coordinates. It then recovers a protein-relative RNA pose through rigid placement guided by these predictions, assembling the complex while preserving the generated RNA conformation.

We evaluate RiboBridge on a temporally split, sequence-filtered, and redundancy-controlled benchmark of 124 protein–RNA complexes. Under post-hoc interface-correspondence evaluation, all RiboBridge candidates are generated before native interfaces are used to select the corresponding predicted-pocket candidate for comparison. With each compared method using its supported conditioning, RiboBridge achieves the highest fNAT and DockQ~\citep{mirabello2024dockq} and the lowest interface and ligand RMSDs. Stage-wise diagnostics identify RNA conformational generation and interaction prediction as the principal sources of complex-recovery error.

Our main contributions are as follows:
\begin{itemize}[
    leftmargin=*,
    itemsep=0pt,
    topsep=1pt,
    parsep=0pt,
    partopsep=0pt
]
    \item We formulate protein-conditioned RNA structure generation in independent RNA and protein coordinate frames, separating conformational generation from relative-pose recovery while retaining predicted-interface conditioning.
    
    \item We introduce \textbf{RiboBridge}, which predicts bound-state interaction constraints from independently represented protein and RNA structures and connects a pretrained RNA structural space to complex assembly through interaction-guided rigid placement.
    
    \item We construct a temporally split, sequence-filtered, and redundancy-controlled benchmark, and combine controlled component ablations with stage-wise diagnostics to assess interface conditioning and localize the main sources of complex-recovery error.
\end{itemize}
\section{Related Work}
\label{sec:related_work}

\subsection{Sequence-based RNA binder design}

Early computational approaches to RNA binder design focused on generating or optimizing RNA sequences from experimental selection data. 
RaptGen and AptaDiff expand aptamer sequence space through latent-variable generation and diffusion-based modeling of selected libraries, respectively~\citep{iwano2022generative,wang2024aptadiff}. Recent approaches further incorporate target protein information for conditional RNA sequence generation, including RNAtranslator and RNA-BAnG~\citep{shukueian2025rnatranslator,roman2025bang}. 
NACraft incorporates structure prediction feedback during sequence optimization to improve candidate refinement~\citep{zhu2026nacraft}. 
These approaches primarily treat sequence as the design variable, while the bound RNA conformation and protein-relative geometry are obtained through downstream prediction or optimization. 
RiboBridge instead focuses on directly generating protein-conditioned RNA structures and modeling their assembly with target proteins.

\subsection{Structure generation and protein--RNA complex modeling}

Recent advances in RNA structure generation have enabled direct modeling of RNA geometries beyond sequence optimization. 
Representative approaches include scaffold-based inverse folding, RNA backbone generation, and joint sequence--structure generation with geometric models such as RhoDesign, RNA-FrameFlow, and RiboFlow~\citep{wong2024deep,anand2025rna,ma2026riboflow}. 
Protein--RNA generation methods further incorporate target protein information through joint modeling of nucleic-acid structures and protein contexts, including MMDiff, RNAFlow, RNA-EFM, RFDpoly, and ODesign~\citep{morehead2023towards,nori2024rnaflow,abir2025rnaefm,favor2025novo,zhang2025odesign}. 
However, these approaches generally model RNA generation and protein-relative placement within a unified complex representation. 
RiboBridge separates RNA conformational generation from protein--RNA assembly by generating RNA structures in an RNA-centric structural space and using cross-frame interaction geometry to connect them with target proteins. 
The pretrained RNA representation from RiboSphere provides the structural prior for this generation process~\citep{zhang2026ribosphere}.
\section{Method}
\label{sec:method}

\begin{figure*}[!htbp]
    \centering
    \includegraphics[width=0.95\linewidth]{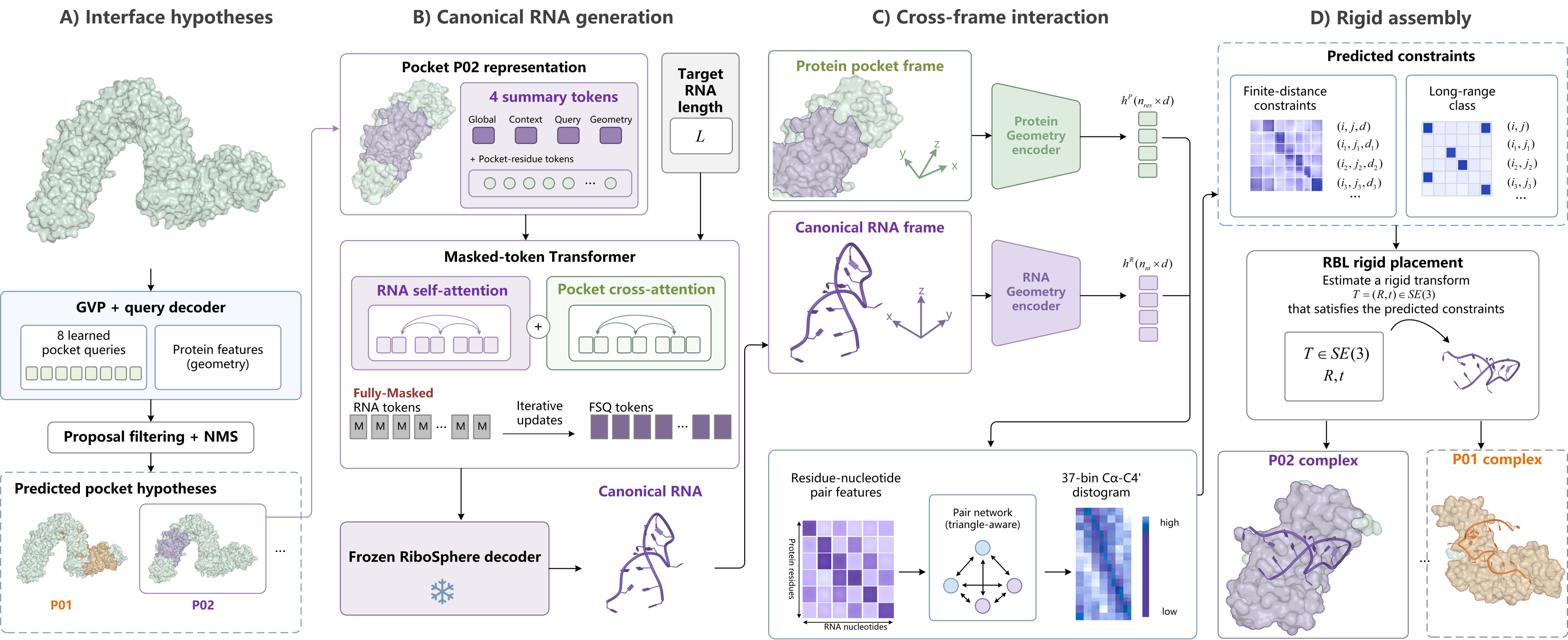}
    \vspace{-2mm}
    \caption{
            \textbf{Overview of the four-stage RiboBridge framework.}
            \textbf{(A)} Interface hypothesis generation.
            \textbf{(B)} Pocket-conditioned canonical RNA generation.
            \textbf{(C)} Cross-frame interaction prediction.
            \textbf{(D)} Interaction-guided rigid placement.
            }
    \label{fig:method_framework}
\end{figure*}

\subsection{Problem formulation}
\label{sec:method_overview}

Given a target protein structure $\mathcal P$ and desired RNA length $L$, RiboBridge generates candidate protein--RNA complexes without a specified binding site or native RNA sequence or structure.
It decomposes the task into four stages: interface hypothesis generation, pocket-conditioned canonical RNA generation, cross-frame interaction prediction, and rigid placement (Figure~\ref{fig:method_framework}).

The interface predictor proposes $M$ candidate regions $\{\Pi_m\}_{m=1}^{M}$,
where $\Pi_m=(\mathcal S_m,\boldsymbol\mu_m,\mathbf Q_m)$
contains a pocket residue set $\mathcal S_m$,
center $\boldsymbol\mu_m\in\mathbb R^3$, and 
local-to-global frame $\mathbf Q_m\in\mathrm{SO}(3)$.
A pocket representation $\mathbf H_m$ conditions a length-$L$ discrete RNA token sequence $\mathbf z_m$,
which the frozen RiboSphere decoder converts into canonical atom11 coordinates 
$\mathbf X_m^{\mathrm{can}}\in\mathbb R^{L\times11\times3}$.
Here, \emph{canonical} denotes an RNA structure without a specified pose relative to the protein; structural representations and coordinate-frame conventions are detailed in Appendix~\ref{app:representations}.

RiboBridge then predicts bound-state protein--RNA distance distributions and recovers a pocket-local rigid transform $(\mathbf R_m,\mathbf t_m)$. The final coordinate of atom $k$ in nucleotide $i$ is

\begin{equation}
    \widehat{\mathbf x}_{mik}^{\mathrm{complex}}
    =
    \boldsymbol\mu_m+
    \mathbf Q_m
    \left(
      \mathbf R_m\mathbf x_{mik}^{\mathrm{can}}+\mathbf t_m
    \right)
    .
    \label{eq:complex_composition}
\end{equation}

Here, $i=1,\ldots,L$ and $k=1,\ldots,11$, and the same rigid transform is applied to all RNA atoms. 
The procedure is repeated independently for each retained interface hypothesis.

\subsection{Interface hypothesis generation}
\label{sec:method_pockets}

Because the binding interface is unknown at inference, RiboBridge predicts multiple interface hypotheses over the full protein surface.

A protein geometry encoder maps residue identities, atom14 coordinates, and local structural environments to rotation-invariant residue representations $\mathbf h_j^P$. 
Learned queries cross-attend to these representations, with the binding probability for query $m$ and residue $j$ defined as

\begin{equation}
\begin{gathered}
    p_{mj}
    =
    \sigma\!\left(
      s\,\overline{\mathbf q}_m^{\top}\overline{\mathbf h}_j^P
      +\alpha\,g(\mathbf a_j;\widetilde{\boldsymbol\mu}_m,\widetilde r_m)
    \right), \\[2pt]
    g(\mathbf a;\boldsymbol\mu,r)
    =
    -\frac{\|\mathbf a-\boldsymbol\mu\|_2^2}{2r^2}.
\end{gathered}
\label{eq:pocket_probability}
\end{equation}

where $\mathbf a_j$ is the C$\alpha$ coordinate, bars denote normalized projections, and $s>0$ is a similarity scale. 
The query-specific center and positive radius are denoted by $\widetilde{\boldsymbol\mu}_m$ and $\widetilde r_m$. 
Query-level matching encourages different queries to capture distinct binding regions ~\citep{carion2020end}.

At inference time, each query yields a compact residue support. 
Residue probabilities determine the interface center, spatial extent, and confidence. 
Pocket-level non-maximum suppression removes lower-confidence proposals that overlap with higher-confidence ones. 
If all proposals fail filtering, we retain the highest-confidence proposal. 
No user-provided seed residue is required.

For each retained proposal, a condition adapter projects global protein context, the query representation, interface geometry, and selected residue features into $\mathbf H_m\in\mathbb R^{N_m\times D}$. 
Each residue feature combines its protein representation, binding score, and pocket-local position $\mathbf Q_m^\top(\mathbf a_j-\boldsymbol\mu_m)$. 
Thus, RNA conformation is conditioned on the predicted interface hypothesis as well as the requested length. 
Architecture, proposal filtering, and condition construction are detailed in Appendix~\ref{app:pocket_details}.

\subsection{Pocket-conditioned canonical RNA generation}
\label{sec:method_canonical_generation}

RNA structural generation should not require learning arbitrary protein-relative orientations from limited protein--RNA complexes. 
RiboBridge therefore generates RNA conformations in an RNA-centric structural space and uses protein information only as conditioning.

High-quality protein--RNA complexes are much scarcer than RNA-only structures. 
Learning atom-level geometry directly from paired data would require relearning generic RNA stereochemistry from this limited set. 
We instead generate discrete tokens in the frozen structural space of RiboSphere. 
Each nucleotide is represented by an FSQ ~\citep{mentzer2024finite} code 
$\mathbf z_i=(z_{i1},\ldots,z_{iD_z})$. 
These codes describe RNA geometry without specifying a protein-relative pose.

During training, entire nucleotide tokens are replaced with MASK at a probability that increases with time. 
Every nonempty RNA retains at least one masked valid position. 
A conditional Transformer reconstructs the masked factors from the corrupted sequence $\widetilde{\mathbf z}_t$, time $t$, and interface condition $\mathbf H_m$. 
Let $\mathcal M_t$ denote the masked valid positions, with $|\mathcal M_t|\geq1$. 
The training objective is

\begin{equation}
\begin{aligned}
\mathcal L_{\mathrm{gen}}
&=
\mathbb E_{t,\mathbf z}
\Bigg[
-\frac{1}{|\mathcal M_t|}
\sum_{i\in\mathcal M_t}
\frac{1}{D_z}
\sum_{d=1}^{D_z}
\log p_\theta
\left(
z_{id}\mid\widetilde{\mathbf z}_t,
\mathbf H_m,t
\right)
\Bigg]
\\
&\quad+
\lambda_{\mathrm{code}}\mathcal L_{\mathrm{code}}
+\lambda_{\mathrm{sc}}\mathcal L_{\mathrm{sc}} .
\end{aligned}
\label{eq:canonical_generation_loss}
\end{equation}

Here, $\mathcal L_{\mathrm{code}}$ regularizes the reconstructed FSQ embedding, while $\mathcal L_{\mathrm{sc}}$ supervises model-generated intermediate states to reduce the train--inference state mismatch.

Inference starts from an all-MASK sequence and iteratively refines nucleotide predictions.
Every position is reconsidered at each step, so early token assignments remain revisable. 
The final tokens are decoded into coordinates by the frozen RiboSphere decoder,

\[
    \mathbf X_m^{\mathrm{can}}
    =
    \mathcal D_{\mathrm{RS}}(\mathbf z_m).
\]

yielding centered RNA geometry.
The masking schedule, self-correction procedure, iterative decoding, and RiboSphere integration are described in Appendix~\ref{app:generator_details}.

\subsection{Cross-frame interaction prediction}
\label{sec:method_interaction_prediction}

Because the canonical RNA and protein interface are defined in independent coordinate frames, directly predicting cross-chain coordinates would require learning an arbitrary alignment. RiboBridge instead predicts invariant interaction geometry that constrains their relative assembly.

Before placement, the relative rotation and translation between the two frames are unknown; therefore pre-placement cross-chain displacements, directions, or raw distances would leak or arbitrarily specify the pose to be inferred. We instead encode each chain independently using rigid-transform-invariant within-chain geometry, including distances, local-frame displacement projections, local shape descriptors, and radial features. A frame-valid indicator handles degenerate local frames. The invariance argument is provided in Appendix~\ref{app:interaction_details}.

Separate geometry-aware encoders update protein anchors and RNA nucleotides. Their embeddings form a two-dimensional pair representation, which is processed by triangle-reasoning updates to predict a bound-state distogram for every protein C$\alpha$--RNA C4$'$ pair. For supervision, the generated canonical RNA is rigidly aligned to native-bound C4$'$ coordinates using an oracle Kabsch transform, after which protein--RNA distance labels are constructed. The oracle pose and native RNA coordinates are used only for label construction and never as network inputs; the cross-frame supervision construction is detailed in Appendix~\ref{app:representations}. The training objective combines multiclass distance prediction with two class-balanced contact losses that emphasize short-range interactions,

\begin{equation}
    \mathcal L_{\mathrm{int}}
    =
    \mathcal L_{\mathrm{dist}}
    +\frac{\lambda_{\mathrm{ct}}}{2}
      \left(
        \mathcal L_{\mathrm{ct}}^{5\text{\AA}}
        +\mathcal L_{\mathrm{ct}}^{8\text{\AA}}
      \right).
    \label{eq:interaction_loss}
\end{equation}

Local frames, the pair architecture, distance discretization, and class balancing are detailed in Appendix~\ref{app:interaction_details}.

\subsection{Interaction-guided rigid placement}
\label{sec:method_placement}

After interaction geometry has been predicted, the remaining task is to recover a rigid placement consistent with the predicted interaction ranges between the RNA and protein frames.

We express protein anchors in the interface-local frame as 
$\mathbf A_{mj}=\mathbf Q_m^\top(\mathbf a_j-\boldsymbol\mu_m)$ 
and denote the canonical C4$'$ coordinate of nucleotide $i$ by $\mathbf x_{mi}$. 
Finite distogram classes yield target ranges $\widehat d_{mji}$, whereas pairs assigned to the long-range class impose separation constraints. 
For a pose $(\mathbf R,\mathbf t)\in\mathrm{SE}(3)$, the corresponding distance is 
$d_{mji}(\mathbf R,\mathbf t)=\|\mathbf A_{mj}-(\mathbf R\mathbf x_{mi}+\mathbf t)\|_2$. 
To accommodate empty constraint sets, define

\[
    \operatorname{avg}_0(f;\Omega)
    =
    \frac{\sum_{q\in\Omega}f_q}{\max(|\Omega|,1)}.
\]

The placement objective combines robust finite-range residuals with a long-range penalty,

\begin{equation}
    \begin{aligned}
    \mathcal E_m(\mathbf R,\mathbf t)
     =
    \operatorname{avg}_0
    &\left(
      \rho_\delta
      \big(d_{mji}(\mathbf R,\mathbf t)-\widehat d_{mji}\big);
      \Omega_{f,m}
    \right)
    \\
    \quad+\lambda_\infty
    \operatorname{avg}_0
    &\left(
      \left[\operatorname{softplus}
      \big(\tau-d_{mji}(\mathbf R,\mathbf t)\big)\right]^2;
      \Omega_{\infty,m}
    \right).
    \end{aligned}
    \label{eq:placement_objective}
\end{equation}

where $\Omega_{f,m}$ and $\Omega_{\infty,m}$ are the finite-range and long-range pair sets, respectively, $\rho_\delta$ is the Huber penalty, and $\tau=20$\,\AA{} is the maximum finite distance represented by the distogram.

The range-based rigid placement solver (RBL) ~\citep{chepuri2014rigid} first obtains a linear initialization from centered squared-range equations, projects the estimated rotation onto $\mathrm{SO}(3)$, and optimizes Eq.~(\ref{eq:placement_objective}) over the six pose degrees of freedom. 
If the linear system is rank-deficient, a nonlinear fallback directly optimizes distance-interval violations. 
The pose Jacobian then determines whether the solution has local finite-range observability in all six degrees of freedom. 
Placement changes only the global rigid pose of the RNA, leaving its canonical conformation intact. 
The derivation and numerical settings are given in Appendix~\ref{app:placement_details}.
\section{Experiments}

We evaluate RiboBridge from three perspectives. 
First, we compare complex recovery with existing protein--RNA design methods under post-hoc interface-correspondence evaluation, then assess sequence realizability and structural self-consistency using AlphaFold~3~\citep{abramson2024accurate}.
Second, stage-wise diagnostics and controlled ablations examine the contributions of RNA conformational generation, interface conditioning, cross-frame interaction prediction, and rigid placement.
Finally, case studies illustrate distinct binding solutions across predicted pockets and RNA lengths.

\subsection{Experimental setup}
\label{sec:setup}

\noindent\textbf{Dataset.}
We curate 29,624 protein--RNA complex records from the Protein Data Bank (PDB) ~\citep{berman2000protein}, each comprising one protein and one RNA, with resolution $\leq$3.5\,\AA{}, RNA lengths of 5--500 nucleotides, and at least five interface contacts (d$<$6\,\AA{}). By PDB release date, 24,100 complexes released before 2023 form the training set and 1,923 released in January--June 2023 form the validation set. Test candidates released after June 30, 2023 must have maximum protein and RNA sequence identities below $40\%$ and $80\%$, respectively. Within-test redundancy reduction at the same thresholds yields 124 test complexes (Appendix~\ref{app:data_details}).

\noindent\textbf{Baselines.}
Our baselines are RNAFlow~\citep{nori2024rnaflow}, ODesign~\citep{zhang2025odesign}, RFDpoly~\citep{favor2025novo}, and NACraft~\citep{zhu2026nacraft}. These methods cover sequence-conditioned, structure-generation, and protein-conditioned RNA design settings. 
We follow each method's standard generation pipeline. 
For the subsequent analysis of sequence realizability, we use NA-MPNN~\citep{kubaney2025rna} to design sequences for the backbones generated by RFDpoly and RiboBridge. For methods that directly produce RNA sequences, we retain their outputs.

\noindent\textbf{Evaluation metrics.}
We report native interface contact recovery (fNAT), interface RMSD (iRMSD),
ligand RMSD (lRMSD), DockQ, and atomic clashes (ClashScore).
We also report DockQ success, defined as the number of targets achieving
DockQ $\geq0.23$ out of all 124 test targets; metric definitions and failure
handling are detailed in Appendix~\ref{app:structural_metrics}.

\begin{table*}[!h]
  \centering
  \caption{
  Complex generation performance on the test set under method-specific conditioning and post-hoc interface correspondence. 
  Each method follows the conditioning supported by its released pipeline; baseline-specific settings are detailed in Appendix~\ref{app:baseline_details}. RMSD values are reported in \AA{}.
  }
  \label{tab:general_performance}
  \small
  \begin{tabular*}{0.95\linewidth}{@{\extracolsep{\fill}}lcccccc@{}}
  \toprule
  Method & fNAT $\uparrow$ & iRMSD $\downarrow$ & lRMSD $\downarrow$ & DockQ $\uparrow$ & DockQ $\geq0.23$ / 124 $\uparrow$ & ClashScore $\downarrow$ \\
  \midrule
  ODesign & \underline{0.0481} & \underline{6.275} & \underline{37.512} & \underline{0.1156} & \underline{15 (12.10\%)} & \underline{10.23} \\
  RFDpoly & 0.0179 & 6.982 & 38.297 & 0.0883 & 9 (7.26\%) & \textbf{9.66} \\
  RNAFlow & 0.0074 & 7.499 & 47.132 & 0.0693 & 7 (5.65\%) & 18.14 \\
  NACraft & 0.0148 & 7.733 & 44.510 & 0.0737 & 10 (8.06\%) & 35.78 \\
  \midrule
  \textbf{RiboBridge} & \textbf{0.0586} & \textbf{5.356} & \textbf{31.135} & \textbf{0.1308} & \textbf{20 (16.13\%)} & 14.96 \\
  \bottomrule
  \end{tabular*}%
\end{table*}

\subsection{General performance of protein--RNA complex generation}
\label{sec:general_performance}

We evaluate whether separating RNA conformational generation from protein-relative assembly supports complex recovery (Table~\ref{tab:general_performance}).
Because this experiment aims to isolate generation and assembly quality from interface ranking, we evaluate each method using the conditioning supported by its released pipeline (Appendix~\ref{app:baseline_details}). For RiboBridge, all candidates are generated from predicted interface hypotheses before the native interface is consulted.
The native interface is used only afterward to establish a post-hoc correspondence and determine which already generated candidate is evaluated (Appendix~\ref{app:candidate_matching}).

RiboBridge achieves the strongest interface and complex-geometry recovery among the compared methods. 
The improvement spans both native-contact recovery and protein-relative RNA placement, indicating that the benefit is not restricted to a single component of complex geometry. 
RiboBridge also achieves the highest DockQ success count, with 20 of 124 targets (16.13\%) reaching DockQ $\geq0.23$. 
RFDpoly achieves the lowest ClashScore, suggesting that steric quality remains a separate challenge from recovering interaction geometry.

\subsection{Sequence realizability and structural self-consistency}
\label{sec:sequence_realizability}

Native-complex recovery does not establish that generated RNA structures can be realized by valid sequences. 
We therefore assess downstream self-consistency using NA-MPNN-designed sequences for RiboBridge and RFDpoly and direct sequence outputs for other methods. 
AlphaFold~3 predicts each complex from the target protein and RNA sequence without access to the designed coordinates.
The complete sequence-design, AlphaFold~3 inference, and self-consistency
evaluation protocols are detailed in Appendix~\ref{app:af3_protocol}.

RiboBridge matches the highest conformational designability ($8.06\%$) and achieves the second-lowest RNA scRMSD and scLRMSD (Table~\ref{tab:sequence_realizability}). 
These results indicate that sequence realization introduces additional challenges beyond backbone generation, reflecting the combined effects of RNA structure generation, sequence design, and AF3 prediction.

\begin{table*}[!h]
  \centering
  \caption{
  Sequence realizability and AlphaFold~3 self-consistency on the 124-target test set. Rec. denotes sequence recovery.
  Designability is the percentage of all 124 test targets that yield an AF3-valid design with RNA scRMSD below 2\,\AA{}.
  }
  \label{tab:sequence_realizability}
  \small
  \begin{tabular*}{0.95\linewidth}{@{\extracolsep{\fill}}llccccc@{}}
  \toprule
  Method
  & Seq. source
  & Rec. $\uparrow$
  & scRMSD (\AA{}) $\downarrow$
  & scLRMSD (\AA{}) $\downarrow$
  & scTM $\uparrow$
  & Des. (\%) $\uparrow$ \\
  \midrule
  ODesign & Direct output & 0.4619 & 10.76 & \textbf{33.91} & 0.0343 & 6.45 \\
  RFDpoly & NA-MPNN & 0.4655 & 10.90 & 42.76 & 0.0638 & \textbf{8.06} \\
  RNAFlow & Direct output & 0.3781 & \textbf{9.31} & 48.24 & 0.0370 & 6.45 \\
  NACraft & Direct output & \textbf{0.4733} & 10.69 & 41.86 & \underline{0.0682} & \underline{7.26} \\
  \midrule
  \textbf{RiboBridge} & NA-MPNN & \underline{0.4683} & \underline{10.14} & \underline{40.98} & \textbf{0.0783} & \textbf{8.06} \\
  \bottomrule
  \end{tabular*}%
\end{table*}

\begin{table*}[!h]
\centering
\caption{
\textbf{Ablation of pocket-conditioned canonical RNA generation.}
All variants are evaluated on the same 54-target formally matched subset.
The lower block reports target-paired bootstrap 95\% confidence intervals
for the advantage of the full model, with positive values favoring Full
for all metrics.
}
\label{tab:ablation_generator}
\small
\begin{tabular*}{0.95\linewidth}{@{\extracolsep{\fill}}lcccc@{}}
\toprule
Variant
& Atom11 RMSD (\AA{}) $\downarrow$
& iRMSD (\AA{}) $\downarrow$
& fNAT $\uparrow$
& DockQ $\uparrow$ \\
\midrule

\multicolumn{5}{l}{\textit{Absolute performance}} \\
Full (pred. matched)
& \textbf{7.972}
& \textbf{4.134}
& \textbf{0.4584}
& \textbf{0.4691} \\
No self-correction
& 8.054
& 4.163
& 0.4500
& 0.4659 \\
Length-only
& 9.459
& 4.984
& 0.3231
& 0.3622 \\
Shuffled pocket
& 9.420
& 4.798
& 0.3296
& 0.3745 \\

\addlinespace[2pt]
\cmidrule(lr){1-5}
\multicolumn{5}{l}{\textit{Target-paired 95\% CI of Full advantage}} \\

Full vs.\ no self-correction
& {\scriptsize $[-0.441,\,0.649]$}
& {\scriptsize $[-0.158,\,0.214]$}
& {\scriptsize $[-0.036,\,0.058]$}
& {\scriptsize $[-0.019,\,0.026]$} \\

Full vs.\ length-only
& {\scriptsize $[0.377,\,2.586]$}
& {\scriptsize $[0.095,\,1.692]$}
& {\scriptsize $[0.059,\,0.219]$}
& {\scriptsize $[0.051,\,0.167]$} \\

Full vs.\ shuffled pocket
& {\scriptsize $[0.347,\,2.586]$}
& {\scriptsize $[0.076,\,1.321]$}
& {\scriptsize $[0.062,\,0.201]$}
& {\scriptsize $[0.049,\,0.144]$} \\

\bottomrule
\end{tabular*}
\end{table*}

\subsection{Analysis of structural decomposition}
\label{sec:ablations}

We analyze the proposed structural decomposition by separately evaluating the contributions of RNA conformational generation, cross-frame interaction prediction, and rigid placement. 
These experiments identify which stages limit final complex recovery and test whether interface conditioning provides information beyond RNA length or generic conditioning.

\subsubsection{Contribution of individual structural stages}
\label{sec:ablation_individual}

\begin{figure*}[!h]
    \centering
    \includegraphics[width=1.\linewidth]{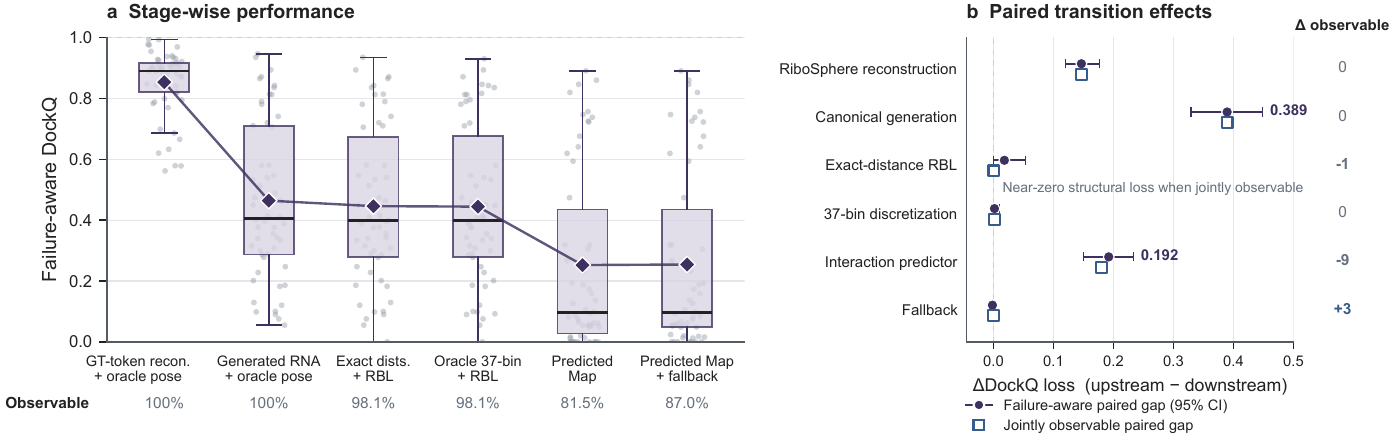}
    \vspace{-2mm}
    \caption{
    \textbf{Error decomposition of RiboBridge.}
    Results are computed on the 54-target diagnostic subset satisfying the
    formal interface-matching criterion.
    \textbf{(a)} Stage-wise failure-aware DockQ and pose observability.
    \textbf{(b)} Paired DockQ changes between consecutive stages. Filled
    markers include unobservable failures; open markers compare targets
    that are observable at both stages.
    }
    \label{fig:ablation_individual}
\end{figure*}

The stage-wise diagnostic is restricted to test targets satisfying the
predefined formal interface-matching criterion in Appendix~\ref{app:candidate_matching}, yielding a
54-target diagnostic subset.
Figure~\ref{fig:ablation_individual} shows that the major performance losses arise from canonical RNA generation and cross-frame interaction prediction. 
Replacing GT-token reconstructions with generated canonical RNA reduces DockQ from 0.854 to 0.464; using predicted interaction maps lowers it further to 0.253. 
By comparison, rigid placement with exact distances introduces minimal additional structural loss. 
These results indicate that improving RNA conformational generation and interaction prediction is more consequential than further refining placement under accurate constraints.

\subsubsection{Effect of interface conditioning on RNA conformational generation}
\label{sec:ablation_generator}

We compare matched predicted-pocket conditioning with length-only and
length-matched shuffled-pocket controls, using canonical-structure metrics
and oracle-pose DockQ to avoid confounding from downstream interaction
prediction and placement.
Matched predicted-pocket conditioning increases oracle-pose DockQ from 0.3622 (length-only) and 0.3745 (shuffled pocket) to 0.4691 (Table~\ref{tab:ablation_generator}). 
For both comparisons, target-paired 95\% bootstrap confidence intervals for the Full advantage lie above zero across all four metrics. 
These results support the contribution of the correctly matched
predicted-pocket condition to RNA conformation generation beyond RNA length
or additional conditioning tokens alone.
By contrast, self-correction yields only small point-estimate gains, with all corresponding intervals including zero.

\begin{table*}[!h]
  \centering
  \caption{
  \textbf{Ablation of cross-frame interaction prediction.}
  Contact F1 and distogram MAE evaluate predictions before rigid placement;
  pose RMSD and DockQ evaluate the resulting complexes.
  CE-only removes the explicit 5\,\AA{} and 8\,\AA{} contact objectives.
  Pose RMSD and DockQ are computed using the fixed reference RBL solver
    without nonlinear fallback.
  }
  \label{tab:ablation_interaction}
  \small
  \begin{tabular*}{0.95\linewidth}{@{\extracolsep{\fill}}lccccc@{}}
  \toprule
  Variant
  & F1@5\,\AA{} $\uparrow$
  & F1@8\,\AA{} $\uparrow$
  & Dist. MAE (\AA{}) $\downarrow$
  & Pose RMSD (\AA{}) $\downarrow$
  & DockQ $\uparrow$ \\
  \midrule
  Full                  & \textbf{0.272} & \textbf{0.388} & \textbf{3.571} & \textbf{21.92} & \textbf{0.253} \\
  $-$ local geometry     & 0.235 & 0.331 & 3.793 & 22.10 & 0.241 \\
  $-$ triangle reasoning & 0.263 & 0.348 & 3.683 & 24.58 & 0.250 \\
  CE-only               & 0.227 & 0.353 & 3.573 & 24.66 & 0.246 \\

  \bottomrule
\end{tabular*}
\end{table*}

\subsubsection{Components of cross-frame interaction prediction}
\label{sec:ablation_interaction}

We ablate local geometry, triangle reasoning, and the explicit 5\,\AA{}/8\,\AA{} contact objectives (CE-only), while holding the canonical RNA inputs and the reference RBL solver without nonlinear fallback fixed. We evaluate contacts and distograms before placement, and pose RMSD and DockQ after placement (Table~\ref{tab:ablation_interaction}). Removing local geometry reduces F1@5\,\AA{} from 0.272 to 0.235 and F1@8\,\AA{} from 0.388 to 0.331, accompanied by the largest DockQ decrease, from 0.253 to 0.241. Removing triangle reasoning has a smaller effect on F1@5\,\AA{} but lowers F1@8\,\AA{} to 0.348 and increases pose RMSD from 21.92 to 24.58\,\AA{} among evaluable targets. CE-only likewise reduces contact F1 and increases pose RMSD to 24.66\,\AA{}, despite nearly unchanged distogram MAE (3.573 versus 3.571\,\AA{}). These results suggest that local geometry, pair reasoning, and explicit contact supervision provide information useful for interaction prediction that is not fully reflected by distogram MAE; the downstream DockQ differences for triangle reasoning and CE-only remain small.

\begin{figure*}[!h]
    \centering
    \includegraphics[width=1.\linewidth]{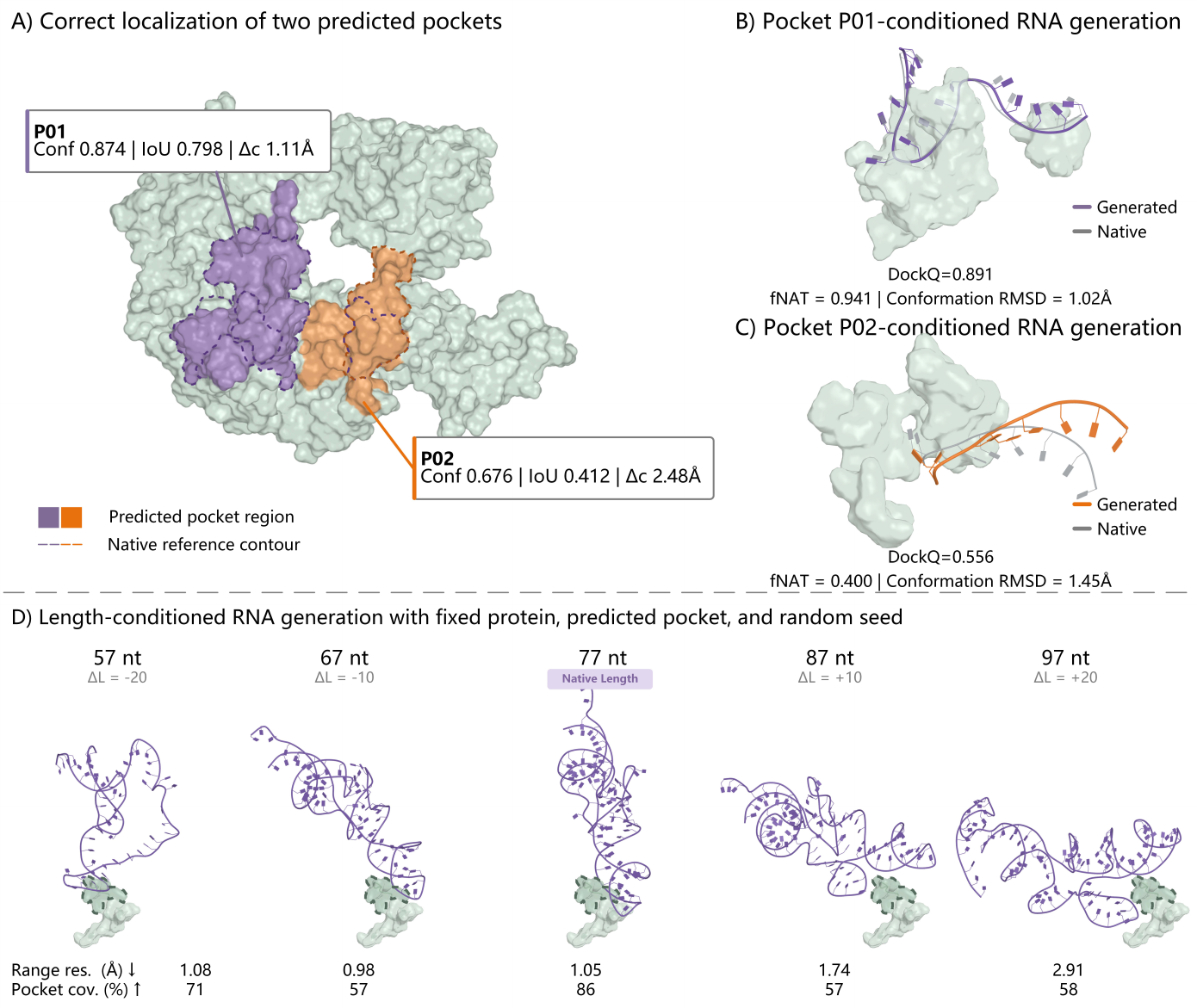}
    \caption{
\textbf{Predicted-pocket-conditioned RNA generation.}
\textbf{(A--C)} Distinct RNA binding solutions generated from two predicted pockets.
\textbf{(D)} Length-conditioned RNA generation at a fixed predicted pocket on a separate target.
}
    \label{fig:casestudy}
\end{figure*}

\subsection{Analysis of generated binding solutions}
\label{sec:cases}

We use two representative examples to analyze distinct binding solutions across predicted interfaces and length-conditioned generation at a fixed predicted pocket (Figure~\ref{fig:casestudy}).

\noindent\textbf{Distinct solutions from predicted interfaces.}
From the protein structure alone, RiboBridge identifies two spatially separated interface hypotheses, each overlapping a corresponding native interface region (Figure~\ref{fig:casestudy}A). Independent generation at the two predicted pockets produces distinct RNA conformations and binding poses, with DockQ values of 0.891 and 0.556, respectively (Figure~\ref{fig:casestudy}B,C). This illustrates that multiple interface hypotheses can yield distinct protein--RNA binding solutions.

\noindent\textbf{Length control at a fixed interface.}
We fix the protein, predicted pocket P01, and random seed while varying the requested RNA length from 57 to 97 nt (Figure~\ref{fig:casestudy}D; $L_{\mathrm{native}}=77$ nt). All generated RNAs attain the requested lengths and contact the same predicted pocket, with pocket coverage of 57--86\%. Range residual remains approximately 1\,\AA{} for 57--77 nt, but increases to 1.74\,\AA{} and 2.91\,\AA{} at 87 and 97 nt, respectively. Thus, RNA length provides a controllable generation condition, while longer outputs in this example become less consistent with the predicted cross-chain distance constraints. Further examples are provided in Appendix~\ref{app:qualitative_recovery}.
\section{Conclusion}

In this paper, we have introduced RiboBridge, a structure-first framework for protein--RNA binder design that separates RNA conformational generation from protein--RNA assembly through cross-frame interaction geometry. 
By generating RNA structures in an RNA-centric structural space and bridging them to protein surfaces through predicted interaction constraints, RiboBridge improves complex structure recovery while providing an interpretable decomposition of the factors governing
successful complex recovery.
A current limitation is that reliable prioritization of binding-interface hypotheses and achieving consistently high-fidelity sequence realization remain challenging for protein--RNA design from structure alone. 
Future work will explore larger-scale functional validation and end-to-end optimization of designed RNA binders by integrating structural generation with sequence and activity objectives.

\section*{Impact Statement}
This work uses publicly available protein--RNA structures and involves
computational modeling rather than human participants or personal data.
The generated RNA structures are predictions of possible binding
configurations; structural recovery does not establish binding affinity,
biological activity, safety, or therapeutic efficacy. Because a general
method for designing protein-binding RNAs could have dual-use
applications, any downstream synthesis or experimental application
should be subject to appropriate institutional biosafety review and
applicable regulations.


\bibliography{example_paper}
\bibliographystyle{icml2026}

\newpage
\appendix
\onecolumn
\section{Supplementary Methods and Analyses}
\label{app:method_details}

\subsection{Data Processing and Experimental Details}
\label{app:data_details}

\paragraph{Dataset construction.}
We curate protein--RNA complex structures from the Protein Data Bank (PDB) following the filtering criteria described in Section~\ref{sec:setup}. 
Structures are retained if the experimental resolution is $\leq 3.5$\,\AA{}, the RNA length is between 5 and 500 nucleotides, and the complex contains at least five protein--RNA interface contacts with distance below $6$\,\AA{}. 
Complexes are split according to PDB release date to reduce temporal information leakage. 
Complexes released before 2023 are used for training, complexes released from January to June 2023 are used for validation, and later structures are reserved for testing.

\paragraph{Sequence filtering and redundancy reduction.}
For test-set construction, candidate complexes released after June 30, 2023 are filtered to satisfy maximum protein and RNA sequence identities of $40\%$ and $80\%$, respectively, relative to the train and valid set. 
Within the test set, redundant complexes are further removed using the same identity thresholds, resulting in a final benchmark of 124 protein--RNA complexes.

\paragraph{Evaluation protocol.}
Unless otherwise specified, evaluation is performed on generated protein--RNA
complexes using the target protein structure as input. For RiboBridge, the
native RNA sequence, native RNA structure, and native interface are not provided
during generation. Under post-hoc interface-correspondence evaluation, native
information is used only after generation to establish correspondence between
already generated predicted-pocket candidates and the native interface, thereby
isolating RNA generation and protein--RNA assembly quality from interface ranking.

\subsection{Structural Representations and Cross-frame Supervision}
\label{app:representations}

RiboBridge represents proteins and RNAs using separate structural coordinate systems. 
Proteins are represented by residue identity, residue number, and up to 14 heavy-atom slots per residue. 
Residues without a resolved C$\alpha$ atom are skipped during parsing. 
RNA structures use the RiboSphere atom11 representation,
\[
    \mathcal A_{11}
    =
    (\mathrm P,\mathrm{C5}^{\prime},\mathrm{C4}^{\prime},
     \mathrm{C3}^{\prime},\mathrm{C2}^{\prime},\mathrm{C1}^{\prime},
     \mathrm{O5}^{\prime},\mathrm{O4}^{\prime},\mathrm{O3}^{\prime},
     \mathrm{O2}^{\prime},\mathrm N_{\mathrm{gly}}).
\]

Here, $\mathrm N_{\mathrm{gly}}$ is N9 for A/G and N1 for C/U. 
Cross-frame interaction prediction and rigid placement use C4$'$ as the nucleotide anchor. 
The final recovered pose is applied to all valid atom11 coordinates.

Equation~(\ref{eq:complex_composition}) relates the RNA canonical frame, interface-local frame, and protein-global frame. 
Interaction supervision is constructed without providing native RNA coordinates or relative poses as network inputs. 
Let $\mathbf x_i^{\mathrm{nat}}$ denote a native RNA C4$'$ coordinate expressed in the corresponding interface-local frame. 
We obtain an oracle rigid transform through proper Kabsch alignment:

\begin{align}
    (\mathbf R_m^\star,\mathbf t_m^\star)
    & =
    \operatorname*{arg\,min}_{\mathbf R\in\mathrm{SO}(3),\,\mathbf t\in\mathbb R^3}
    \sum_{i=1}^{L}
    \left\|
      \mathbf R\mathbf x_{mi}^{\mathrm{can}}
      +
      \mathbf t
      -
      \mathbf x_i^{\mathrm{nat}}
    \right\|_2^2,
    \nonumber\\
    d_{mji}^\star
    & =
    \left\|
      \mathbf A_{mj}
      -
      \left(
        \mathbf R_m^\star\mathbf x_{mi}^{\mathrm{can}}
        +
        \mathbf t_m^\star
      \right)
    \right\|_2 .
    \label{eq:app_oracle_supervision}
\end{align}

Distogram labels are obtained by discretizing $d_{mji}^\star$. 
Native coordinates therefore determine only the rigid transform used to construct supervision; the resulting interaction labels preserve the intrinsic conformation of the generated canonical RNA. 
Neither $(\mathbf R_m^\star,\mathbf t_m^\star)$ nor native RNA coordinates are provided as model inputs.

\subsection{Interface Hypothesis Generation}
\label{app:pocket_details}

\paragraph{Protein geometry encoder.}
The binding interface is not specified during inference. 
RiboBridge therefore maintains multiple interface hypotheses over the protein surface rather than relying on a predefined interaction region.

A protein geometry encoder maps residue identities, atom14 coordinates, and local structural environments to rotation-invariant residue representations $\mathbf h_j^P$. 
Vector channels encode backbone and side-chain directions, while scalar channels describe residue-level structural and chemical properties. 
The protein graph connects each residue to its 30 nearest neighbors. 
Edge scalars comprise radial basis function (RBF) encodings of C$\alpha$ distances and relative sequence positions, and the edge vector is the normalized C$\alpha$ displacement. 
A four-layer geometric vector perceptron (GVP) ~\citep{jing2020learning} encoder produces 768-dimensional residue representations. 
Only scalar channels and vector-channel norms are passed to the proposal head, making these representations invariant to global rotations.

Eight learned queries interact with the protein representations through a three-layer cross-attention decoder. 
Two refinement blocks predict residue membership, interface center, radius, core region, and proposal quality. 
Training uses multi-interface Hungarian matching. 
The matching cost consists primarily of class-balanced binary cross-entropy (BCE) and Dice terms, supplemented by center and radius terms. 
After matching, auxiliary supervision is applied to region geometry, contact cores, subregions, coverage, and proposal quality.

\paragraph{Interface proposal construction.}
For each query, we retain residues with membership probability at least $0.60$. 
If fewer than six residues pass this threshold and enough valid residues are available, we add the highest-probability residues until the support contains six residues. 
Hard supports are capped at 128 residues. 
Let $h_{mj}$ be the resulting binary support and $g_{mj}$ the geometry-head probability. 
The normalized weights, interface center, and root-mean-square radius are

\begin{equation}
    w_{mj}
    =
    \frac{g_{mj}h_{mj}}
    {\sum_k g_{mk}h_{mk}},
    \qquad
    \boldsymbol\mu_m
    =
    \sum_j w_{mj}\mathbf a_j,
    \qquad
    r_m
    =
    \left(
      \sum_j w_{mj}
      \|\mathbf a_j-\boldsymbol\mu_m\|_2^2
    \right)^{1/2}.
    \label{eq:app_pocket_geometry}
\end{equation}

If the geometry weights are degenerate, we use uniform weights over the support. 
A valid interface hypothesis must have quality confidence $\geq0.05$, at least six residues, and $r_m\geq2$\,\AA{}. 
Proposals are sorted by confidence. 
A lower-confidence proposal is suppressed if its Dice overlap with a higher-confidence proposal is $\geq0.60$, or if their center distance is $\leq6$\,\AA{} and Dice overlap is $\geq0.30$. 
Suppressed supports are not merged into retained proposals. 
If filtering and non-maximum suppression leave no valid proposals, we retain the highest-confidence query among all queries.

\paragraph{Interface-local frame.}
We construct an interface-local coordinate frame from the weighted covariance of the selected C$\alpha$ coordinates. 
The eigenvector associated with the smallest eigenvalue defines $\mathbf e_z$ after resolving its sign to point outward. 
The eigenvector associated with the largest eigenvalue is orthogonalized to define $\mathbf e_x$. 
We then set

\begin{equation}
    \mathbf e_y=\mathbf e_z\times\mathbf e_x,
    \qquad
    \mathbf e_x=\mathbf e_y\times\mathbf e_z,
    \qquad
    \mathbf Q_m=[\mathbf e_x\ \mathbf e_y\ \mathbf e_z].
    \label{eq:app_pocket_frame}
\end{equation}

We use the identity frame if fewer than three points are selected or eigendecomposition fails. 
If orthogonalization of $\mathbf e_x$ is numerically degenerate, the fallback uses the sequence direction and then a coordinate axis. 
This principal component analysis construction does not assume that eigenvectors are uniquely defined for nearly symmetric interfaces.

\paragraph{Condition tokens.}
The conditioning sequence consists of four summary tokens for the global protein, interface query, context summary, and interface geometry, followed by at most 64 selected residue tokens. 
If a hard support contains more than 64 residues, we retain the top 64 by interface probability; otherwise, we use the entire support. 
The first three summary representations are projected from 768 dimensions to $D=512$. 
The geometry token encodes normalized radius, proposal confidence, and the support fraction.

Each residue token combines $\mathbf h_j^P$, the interface score, and the scaled interface-local position

\begin{equation}
    \widetilde{\mathbf a}_{mj}
    =
    \mathbf Q_m^\top
    (\mathbf a_j-\boldsymbol\mu_m)/10.
    \label{eq:app_local_residue_position}
\end{equation}

We use neither soft-interface gating nor an additional interface-attention bias in the reported experiments.

\subsection{Canonical RNA Generation}
\label{app:generator_details}

\paragraph{Factorized structural tokens.}
RiboBridge generates RNA conformations in an RNA-centric structural space rather than directly predicting protein-relative coordinates. 
The frozen RiboSphere encoder maps centered atom11 coordinates to discrete representations through finite scalar quantization ~\citep{mentzer2024finite}. 
Each nucleotide has five categorical factors,

\begin{equation}
    \mathbf z_i=(z_{i1},\ldots,z_{i5}),
    \qquad
    z_{id}\in\{0,\ldots,K_d-1\},
    \qquad
    (K_1,\ldots,K_5)=(7,5,5,5,5),
    \label{eq:app_fsq_factors}
\end{equation}

which define 4375 joint codes and a 512-dimensional code embedding. 
These codes describe RNA geometry without specifying a protein-relative pose.

\paragraph{Masked-token training.}
The conditional Transformer has eight blocks, 16 attention heads, a hidden dimension of 512, a feed-forward expansion factor of four, and dropout of $0.1$. 
For $t\sim\mathcal U(0,1)$, an entire nucleotide token is replaced with MASK with probability

\begin{equation}
    \rho(t)
    =
    \sin^2\!\left(\frac{\pi t}{2}\right)
    \label{eq:app_mask_schedule}
\end{equation}

The factorized cross-entropy in Eq.~(\ref{eq:canonical_generation_loss}) is evaluated only at masked valid positions. 
If Bernoulli corruption masks no valid nucleotide, one valid position is selected uniformly at random as a masked target. 
Thus, every nonempty RNA satisfies $|\mathcal M_t|\geq1$. 
The code regularizer uses SmoothL1 with $\lambda_{\mathrm{code}}=0.05$.

After an initial bootstrap phase, the probability of using on-policy samples for self-correction increases linearly to $0.5$. 
Selected samples are rolled out with the current generator to early decoding steps $1$--$4$ or middle steps $11$--$18$. 
Token and code losses are then recomputed on these model-generated states with $\lambda_{\mathrm{sc}}=1$. 
This supervision applies only to discrete generation and introduces no protein-relative coordinate targets. 
If a batch contains no selected on-policy samples, its self-correction term is zero.

\paragraph{Iterative decoding.}
Inference starts from an all-MASK sequence. 
At step $k$, we select the argmax prediction for each factor and define nucleotide confidence as the geometric mean of the five factor probabilities,

\begin{align}
    \widehat z_{id}^{(k)}
    &=
    \operatorname*{arg\,max}_{c}
    p_\theta
    \left(
      z_{id}=c
      \mid
      \mathbf z^{(k-1)},\mathbf H_m,t_k
    \right),
    \nonumber\\
    s_i^{(k)}
    &=
    \exp\!\left[
      \frac{1}{5}
      \sum_{d=1}^{5}
      \log
      p_\theta
      \left(
        \widehat z_{id}^{(k)}
        \mid
        \mathbf z^{(k-1)},\mathbf H_m,t_k
      \right)
    \right].
    \label{eq:app_token_confidence}
\end{align}

After a stable descending sort by $s_i^{(k)}$, we retain

\begin{equation}
    n_k
    =
    \left\lfloor
      L\sin^2\!\left(\frac{\pi k}{2K}\right)
    \right\rfloor,
    \qquad K=20
    \label{eq:app_unmask_schedule}
\end{equation}

positions and reset the remaining positions to MASK. 
All positions are predicted anew at every step, so earlier assignments can be revised. 
Because decoding does not use multinomial sampling, a fixed condition, RNA length, and checkpoint produce the same token sequence.

\paragraph{RiboSphere coordinate decoding.}
The final RNA tokens are converted into canonical coordinates using the frozen RiboSphere decoder,

\[
    \mathbf X_m^{\mathrm{can}}
    =
    \mathcal D_{\mathrm{RS}}(\mathbf z_m).
\]

RiboSphere has a two-layer encoder with hidden dimension 256 and an eight-layer conditional coordinate decoder with hidden dimension 512. 
The decoder starts from centered Gaussian coordinates and performs 20 fixed integration steps. 
Both the classifier-free guidance weight and the score weight are 1, and the per-step noise weight is 1. 
Valid atom11 coordinates are re-centered with a mask after every step. 
Initial coordinates are generated with a deterministic seed, making the canonical structure reproducible for fixed inputs.

\subsection{Cross-frame Interaction Modeling}
\label{app:interaction_details}

The protein interface and canonical RNA occupy independent coordinate frames before placement. 
The purpose of the interaction module is to bridge these two structural spaces by predicting geometry that constrains their relative assembly.

For a node with a valid orthonormal local frame $\mathbf E_i$, let
$\mathbf r_{ij}=\mathbf x_j-\mathbf x_i$.
Under a rigid transform $(\mathbf S,\mathbf b)\in\mathrm{SE}(3)$, the local direction satisfies

\begin{equation}
    (\mathbf S\mathbf E_i)^\top
    \frac{\mathbf S(\mathbf x_j-\mathbf x_i)}
         {\|\mathbf S(\mathbf x_j-\mathbf x_i)\|_2}
    =
    \mathbf E_i^\top
    \frac{\mathbf x_j-\mathbf x_i}
         {\|\mathbf x_j-\mathbf x_i\|_2}.
    \label{eq:app_independent_frame_invariance}
\end{equation}

Translation cancels in the displacement, while distances, covariance eigenvalue ratios, and local bend are also invariant. 
These geometric features therefore allow each chain to undergo an independent rigid transformation without constructing pre-placement cross-chain coordinates.

\paragraph{Node geometry.}

Protein anchors consist of the selected interface core and up to 16 additional context C$\alpha$ atoms nearest to that core. 
A protein pseudo-frame is constructed from ordered anchor triplets. 
For neighboring normalized displacements $\mathbf v_-$ and $\mathbf v_+$, define

\begin{equation}
    \mathbf e_1
    =
    \operatorname{norm}(\mathbf v_-+\mathbf v_+),
    \qquad
    \mathbf e_3
    =
    \operatorname{norm}(\mathbf v_-\times\mathbf v_+),
    \qquad
    \mathbf e_2
    =
    \operatorname{norm}(\mathbf e_3\times\mathbf e_1).
    \label{eq:app_protein_frame}
\end{equation}

This is an \emph{ordered-anchor pseudo-frame}, not a frame based strictly on sequence-backbone adjacency. 
Terminal, padded, or collinear triplets receive a zero frame and are marked by a frame-valid indicator. 
These degenerate nodes are not assigned an orthonormal frame.

The RNA frame is centered at C4$'$. 
Its first axis follows the C3$'$--C4$'$ direction; its second axis is obtained by orthogonalizing the C1$'$--C4$'$ direction against the first. 
Missing atoms or degenerate geometry trigger a C4$'$ pseudo-frame fallback.

Each protein or RNA node also has a six-dimensional shape descriptor containing the mean and standard deviation of eight-nearest-neighbor distances, two covariance eigenvalue ratios, local bend, and a frame-valid indicator.

Within-chain edge features concatenate normalized displacement directions expressed in both endpoint frames, distance, and frame-valid indicators. 
RNA node features combine a 512-dimensional RiboSphere embedding, a 32-dimensional FSQ embedding, normalized sequence position, and a radial feature. 
Protein node features combine a 512-dimensional interface conditioning feature with interface-centered radius.

\paragraph{Pair network.}

Each chain encoder has four geometry-aware message-passing layers. 
Pairwise concatenation of the resulting 128-dimensional node embeddings forms an $N_A\times L$ pair grid, where $N_A$ is the number of interface anchors. 
The grid is processed by 20 triangle reasoning stages. 
Each stage first applies $1\times9$, $3\times3$, and $9\times1$ residual operations, then performs triangle aggregation using internal protein--protein and RNA--RNA distances. 
No cross-chain coordinates are used.

\paragraph{Distogram supervision.}

Protein C$\alpha$--RNA C4$'$ distances are discretized into 37 classes,

\begin{equation}
    y(d)
    =
    \begin{cases}
      0, & d<2.5,\\
      b, & d\in[2+0.5b,\,2.5+0.5b),
      \quad b=1,\ldots,34,\\
      35, & d\in[19.5,20],\\
      36, & d>20 .
    \end{cases}
    \label{eq:app_distogram_bins}
\end{equation}

with distances measured in \AA{}. 
Class 36 represents long-range pairs.

For strict contact thresholds, we use

\begin{equation}
    p_{ji}^{<5}
    =
    \sum_{b=0}^{5}p_{ji}^{(b)},
    \qquad
    p_{ji}^{<8}
    =
    \sum_{b=0}^{11}p_{ji}^{(b)},
    \qquad
    y_{ji}^{<u}
    =
    \mathbb 1[d_{ji}^{\star}<u].
    \label{eq:app_contact_probabilities}
\end{equation}

Thus, distances exactly equal to $5$\,\AA{} or $8$\,\AA{} are excluded from their respective contact classes.

Let $\Omega$ be the set of valid, unpadded protein--RNA pairs. 
Positive and negative class weights are estimated from training pairs only,

\begin{equation}
    w_+^u
    =
    \frac{1/2}{N_+^u/(N_+^u+N_-^u)},
    \qquad
    w_-^u
    =
    \frac{1/2}{N_-^u/(N_+^u+N_-^u)}.
    \label{eq:app_contact_weights}
\end{equation}

The balanced binary negative log-likelihood is

\begin{equation}
    \mathcal L_{\mathrm{ct}}^u
    =
    -
    \mathbb E_{(j,i)\in\Omega}
    \left[
      w_+^u y_{ji}^{<u}\log p_{ji}^{<u}
      +
      w_-^u(1-y_{ji}^{<u})
      \log(1-p_{ji}^{<u})
    \right].
    \label{eq:app_contact_loss}
\end{equation}

Threshold probabilities are computed stably from logits using log-sum-exp.

\subsection{Interaction-guided Rigid Placement}
\label{app:placement_details}

After interaction geometry has been predicted, the remaining task is to recover a physically consistent relative placement between the independently generated RNA conformation and the protein interface.

\paragraph{Ranges from distograms.}

For each pair, we select the most probable distance class. 
Classes $0$--$35$ are converted to finite reference ranges,

\begin{equation}
    \widehat d(b)
    =
    \begin{cases}
      1.25, & b=0,\\
      2.25+0.5b, & 1\leq b\leq35 ,
    \end{cases}
    \label{eq:app_range_midpoints}
\end{equation}

corresponding to $1.25,2.75,\ldots,19.75$\,\AA{}. 
Class 36 contributes only a long-range constraint.

\paragraph{Linear initialization.}

Let $\overline{\mathbf A}$ and $\overline{\mathbf X}$ be the unweighted arithmetic means of the protein and RNA anchors, respectively. 
Define centered coordinates

\[
\mathbf a_j=\mathbf A_j-\overline{\mathbf A},
\qquad
\mathbf x_i=\mathbf X_i-\overline{\mathbf X},
\]

and denote the centered translation by $\mathbf u$.

For finite ranges associated with RNA anchor $i$, define

\begin{equation}
    q_{ji}
    =
    \frac{\|\mathbf a_j\|_2^2-\widehat d_{ji}^{\,2}}{2}.
    \label{eq:app_centered_range_term}
\end{equation}

Let $\mathcal J_i$ contain the protein anchors with a finite range to RNA anchor $i$. 
For nonnegative weights $w_{ji}$, define

\begin{equation}
    \overline{\mathbf a}_{i,w}
    =
    \frac{\sum_{j\in\mathcal J_i}w_{ji}\mathbf a_j}
         {\sum_{j\in\mathcal J_i}w_{ji}},
    \qquad
    \overline q_{i,w}
    =
    \frac{\sum_{j\in\mathcal J_i}w_{ji}q_{ji}}
         {\sum_{j\in\mathcal J_i}w_{ji}}.
    \label{eq:app_weighted_means}
\end{equation}

Weighted centering of the squared-range equations over corresponding protein anchors eliminates $\|\mathbf R\mathbf x_i+\mathbf u\|_2^2$, yielding

\begin{equation}
    (\mathbf a_j-\overline{\mathbf a}_{i,w})^\top
    (\mathbf R\mathbf x_i+\mathbf u)
    =
    q_{ji}-\overline q_{i,w}.
    \label{eq:app_centered_range_equation}
\end{equation}

Only RNA anchors with at least four finite ranges are used. 
Multiplying each row by $\sqrt{w_{ji}}$ and stacking gives

\begin{equation}
    \mathbf B
    \begin{bmatrix}
      \operatorname{vec}_{\mathrm{row}}(\mathbf R)\\
      \mathbf u
    \end{bmatrix}
    =
    \mathbf b,
    \qquad
    \mathbf B\in\mathbb R^{K_r\times12}.
    \label{eq:app_linear_pose_system}
\end{equation}

When $\operatorname{rank}(\mathbf B)\geq12$, the least-squares rotation estimate is projected onto $\mathrm{SO}(3)$,

\begin{equation}
    \mathbf R_0
    =
    \mathbf U
    \operatorname{diag}
    \left(
      1,1,\det(\mathbf U\mathbf V^\top)
    \right)
    \mathbf V^\top .
    \label{eq:app_rotation_projection}
\end{equation}

With $\mathbf R_0$ fixed, we recover translation

\begin{equation}
    \mathbf t_0
    =
    \overline{\mathbf A}
    +\mathbf u
    -\mathbf R_0\overline{\mathbf X}.
    \label{eq:app_translation_recovery}
\end{equation}

The initial least-squares solve uses $w_{ji}=1$, followed by three rounds of Huber iteratively reweighted least squares (IRLS). 
The finite-range weights are updated as

\begin{equation}
    w_{ji}
    =
    \begin{cases}
      1, & |r_{ji}|\leq\delta_{\mathrm{IRLS}},\\
      \delta_{\mathrm{IRLS}}/|r_{ji}|, & |r_{ji}|>\delta_{\mathrm{IRLS}},
    \end{cases}
    \qquad
    \delta_{\mathrm{IRLS}}=2 .
    \label{eq:app_irls_weights}
\end{equation}

\paragraph{Pose refinement.}

Rotation is parameterized as

\[
\mathbf R(\boldsymbol\omega)
=
\exp([\boldsymbol\omega]_\times)\mathbf R_0 .
\]

We optimize Eq.~(\ref{eq:placement_objective}). 
Finite-range residuals use a Huber penalty with $\delta=1$, and the long-range coefficient is $0.1$. 
Adam runs for at most 300 steps, with rotation and translation learning rates of $0.01$ and $0.05$, respectively. 
The gradient norm is clipped at $10$, and early-stopping patience is 100 steps.

\paragraph{Nonlinear fallback.}

If the linear system fails the rank condition, each finite distance class is interpreted as an interval $[l_{ji},u_{ji}]$ with violation

\begin{equation}
    v_{ji}(d)
    =
    \max(l_{ji}-d,0)+\max(d-u_{ji},0).
    \label{eq:app_interval_violation}
\end{equation}

We initialize the solver with the identity rotation and 24 proper signed permutations constructed from the protein and RNA PCA frames. 
The identity rotation is paired with a centroid translation weighted by finite-range counts. 
All starts are ranked by the interval objective, and the best eight are refined.

The fallback objective is

\begin{equation}
    \mathcal E_{\mathrm{fb}}
    =
    \operatorname{avg}_0
    \left(
      \operatorname{SmoothL1}(v_{ji}(d_{ji}),0);
      \Omega_f
    \right)
    +
    0.1
    \operatorname{avg}_0
    \left(
      [\operatorname{softplus}(20-d_{ji})]^2;
      \Omega_\infty
    \right).
    \label{eq:app_fallback_objective}
\end{equation}

Let $\mathbf r_i=\mathbf R\mathbf x_i$, $\mathbf y_i=\mathbf r_i+\mathbf t$, and

\[
\mathbf n_{ji}
=
\frac{\mathbf y_i-\mathbf A_j}
{\|\mathbf y_i-\mathbf A_j\|_2}.
\]

Under a left-multiplicative rotation perturbation, the six-dimensional pose Jacobian row is

\begin{equation}
    \mathbf J_{ji}
    =
    [
      \mathbf r_i\times\mathbf n_{ji},
      \mathbf n_{ji}
    ].
    \label{eq:app_pose_jacobian}
\end{equation}

A pose is accepted only if its coordinates are finite, $\det(\mathbf R)>0$, and $\operatorname{rank}(\mathbf J)=6$. 
This rank test checks whether the finite-range geometry locally constrains all six motion degrees of freedom at the recovered pose. 
It does not prove global uniqueness of the solution. 
Long-range constraints are excluded from the Jacobian rank test.

\subsection{Architecture and Inference Settings}
\label{app:hyperparameters}

\begin{table}[!h]
    \centering
    \small
    \begin{tabular}{@{}p{0.65\columnwidth}r@{}}
        \hline
        Setting & Value \\
        \hline
        Protein encoder width / GVP layers & $768 / 4$ \\
        Protein graph neighbors & $30$ \\
        Interface queries / query-decoder layers & $8 / 3$ \\
        Proposal refinement layers & $2$ \\
        Interface probability threshold & $0.60$ \\
        Interface support size & $6$--$128$ \\
        Proposal confidence / radius thresholds & $0.05 / 2$\,\AA{} \\
        Interface NMS Dice threshold & $0.60$ \\
        Interface NMS center / relaxed Dice & $6$\,\AA{} / $0.30$ \\
        Condition dimension / summary tokens & $512 / 4$ \\
        Maximum residue tokens & $64$ \\
        Conditional Transformer blocks / heads & $8 / 16$ \\
        Transformer width / FF multiplier / dropout & $512 / 4 / 0.1$ \\
        FSQ levels / joint codes & $(7,5,5,5,5) / 4375$ \\
        Code / self-correction weights & $0.05 / 1.0$ \\
        Maximum on-policy probability & $0.5$ \\
        Token decoding / coordinate decoding steps & $20 / 20$ \\
        \hline
    \end{tabular}
    \caption{
    Architecture and inference settings for interface hypothesis generation and canonical RNA generation.
    }
    \label{tab:app_generation_hyperparameters}
\end{table}

\begin{table}[!h]
    \centering
    \small
    \begin{tabular}{@{}p{0.65\columnwidth}r@{}}
        \hline
        Setting & Value \\
        \hline
        Per-chain hidden dimension & $128$ \\
        Per-chain message-passing layers & $4$ \\
        Triangle reasoning stages & $20$ \\
        Protein context anchors & at most $16$ \\
        Local shape neighborhood size & $8$ \\
        Distogram classes / maximum finite range & $37 / 20$\,\AA{} \\
        Contact thresholds / loss coefficient & $5,8$\,\AA{} / $1$ \\
        Linear-system rank threshold & $12$ \\
        IRLS rounds / Huber parameter & $3 / 2$ \\
        Pose-refinement steps & at most $300$ \\
        Rotation / translation learning rates & $0.01 / 0.05$ \\
        Gradient clip / patience & $10 / 100$ \\
        Fallback starts / refined starts & $25 / 8$ \\
        Pose-Jacobian rank threshold & $6$ \\
        \hline
    \end{tabular}
    \caption{
    Architecture and inference settings for cross-frame interaction modeling and rigid placement.
    }
    \label{tab:app_placement_hyperparameters}
\end{table}

\section{Evaluation Protocol}
\label{app:evaluation_protocol}

This section provides additional details on candidate generation, interface assignment, baseline conditioning, structural evaluation, and AlphaFold~3  self-consistency analysis used in the main experiments.
These protocols are designed to separate three factors that are otherwise coupled in protein-conditioned RNA design: interface identification, RNA conformational generation, and protein--RNA assembly.
Unless otherwise specified, structural evaluations are conducted on the 124-complex test set described in the main paper.

\subsection{Candidate Generation and post-hoc interface correspondence}
\label{app:candidate_matching}

\paragraph{Interface hypothesis generation.}
Given a full target protein, RiboBridge first generates multiple candidate interface hypotheses without access to the native RNA or native protein--RNA interface.
The pocket predictor contains eight learned queries, each of which predicts a residue-level interface probability over the target protein.
At inference time, residues with predicted probability at least $0.60$ are retained.
If fewer than six residues satisfy this threshold, the highest-probability valid residues are added until the support contains six residues.
Each support is capped at 128 residues.
The interface center and spatial extent are then computed from the predicted residue weights.
A proposal is retained only if its confidence is at least $0.05$, its support contains at least six residues, and its radius is at least 2\,\AA{}.

To remove redundant queries that identify the same local region, proposals are processed in descending confidence order using non-maximum suppression.
For a proposal $p$ and a previously retained proposal $p'$, the lower-confidence proposal is suppressed if
\begin{equation}
    \operatorname{Dice}(p,p') \geq 0.60,
\end{equation}
or if
\begin{equation}
    \|\boldsymbol{\mu}_{p}-\boldsymbol{\mu}_{p'}\|_2
    \leq 6\,\text{\AA},
    \qquad
    \operatorname{Dice}(p,p') \geq 0.30.
\end{equation}
If no proposal survives filtering, the globally highest-confidence query is retained.
This procedure depends only on the target protein and the pocket predictor and does not use native protein--RNA interaction information.

\paragraph{Structure generation for each interface hypothesis.}
Each retained interface hypothesis is processed independently by the remaining stages of RiboBridge.
Given the interface condition and requested RNA length, the canonical generator first produces one RNA conformation.
The cross-frame interaction module then predicts protein--RNA interaction geometry, and RBL recovers the corresponding rigid-body pose.
Each retained pocket therefore gives rise to one complete protein--RNA candidate.

The main experiments do not generate a best-of-$K$ stochastic ensemble for each pocket and do not use an additional learned cross-pocket ranker.
Discrete structural tokens are obtained by deterministic argmax-based progressive decoding.
For a fixed checkpoint, interface condition, and requested RNA length, this procedure therefore produces a deterministic canonical RNA candidate.

\paragraph{Post-hoc interface-correspondence evaluation on the 124-target benchmark.}
The main comparison is designed to evaluate RNA conformational generation and protein--RNA assembly given a corresponding interface hypothesis, rather than the ranking of multiple predicted interfaces.
All predicted interface hypotheses are therefore generated before the native interface is consulted.
Native information is used only afterward to establish the post-hoc
evaluation correspondence between an already generated predicted-interface
candidate and the native binding region.

Let
$\{\widehat{\Pi}_{m}\}_{m=1}^{M}$
denote the retained predicted interfaces for a target, and let $\Pi^\star$ denote the native evaluation region.
For each predicted interface $\widehat{\Pi}_{m}$, we compute
\begin{equation}
    D_m =
    \frac{
        2|\widehat{\Pi}_{m}\cap\Pi^\star|
    }{
        |\widehat{\Pi}_{m}|+|\Pi^\star|
    },
    \qquad
    \delta_m =
    \|\widehat{\boldsymbol{\mu}}_{m}
      -\boldsymbol{\mu}^{\star}\|_2 .
    \label{eq:app_interface_match_score}
\end{equation}
The predicted hypothesis with the largest Dice overlap $D_m$ is selected for evaluation.
Ties are resolved by preferring the hypothesis with the smaller center distance $\delta_m$.
If both quantities are identical, the original retained-pocket order provides a deterministic final tie break.

This assignment only determines which already generated predicted-interface candidate is compared with the native complex.
The native interface does not enter pocket prediction, canonical RNA generation, cross-frame interaction prediction, or rigid placement.
The protocol therefore evaluates RNA generation and assembly at the retained predicted interface with the greatest native overlap, without requiring model-based ranking of the retained candidates.

\paragraph{Strict matching for stage-wise analysis.}
The stage-wise error decomposition in the main paper uses a stricter predefined interface-matching criterion.
A predicted interface $\widehat{\Pi}_{m}$ is considered formally matched to the native region $\Pi^\star$ only when
\begin{equation}
    D_m \geq 0.30
    \qquad\text{and}\qquad
    \delta_m \leq 6\,\text{\AA}.
    \label{eq:app_formal_interface_match}
\end{equation}
When multiple native regions or multiple compatible queries are present, a deterministic maximum-cardinality bipartite matching is used to establish one-to-one assignments.
Candidate edges are prioritized by larger Dice overlap, smaller center distance, higher proposal confidence, and smaller query index.

Applying this criterion to the test set yields 54 targets with a formal predicted-pocket match.
This 54-target cohort is used only for stage-wise error decomposition and component-level diagnostics.
It is distinct from the 124-target evaluation-time correspondence used in the main comparison.

\subsection{Definition of Native Interfaces and Baseline Conditioning}
\label{app:native_interface_definition}

\paragraph{Native contact core.}
We construct a protein-side interaction region from each native protein--RNA complex.
For protein residue $j$, let $\mathcal A_j^{P}$ denote its valid atom14 coordinates and let $\mathcal A^{R}$ denote all valid RNA atom11 coordinates.
Residue $j$ is assigned to the native contact core $\mathcal C^\star$ if
\begin{equation}
    \min_{\substack{
        a\in\mathcal A_j^{P}\\
        b\in\mathcal A^{R}
    }}
    \|\mathbf x_a-\mathbf x_b\|_2
    < 6\,\text{\AA}.
    \label{eq:app_native_contact_core}
\end{equation}
The RNA atom11 representation contains
P, C5$'$, C4$'$, C3$'$, C2$'$, C1$'$, O5$'$, O4$'$, O3$'$, O2$'$,
and the glycosidic nitrogen N9/N1.
Only protein residues with valid C$\alpha$ coordinates are considered when constructing the subsequent pocket representation.
If no residue satisfies Eq.~\ref{eq:app_native_contact_core}, the closest valid protein residue is used as the contact-core seed.

\paragraph{Expanded native interface.}
The raw contact core contains only residues in direct atomic contact with the RNA and does not fully describe the surrounding pocket environment.
We therefore construct an expanded native interface
\begin{equation}
    \Pi^\star =
    \left\{
    j:
    \min_{k\in\mathcal C^\star}
    \|
        \mathbf a_j-\mathbf a_k
    \|_2
    \leq 8\,\text{\AA}
    \right\},
    \label{eq:app_expanded_native_interface}
\end{equation}
where $\mathbf a_j$ denotes the C$\alpha$ coordinate of residue $j$.
If the resulting region contains fewer than six residues, additional residues are included according to their distance to the contact core.
The predicted-pocket correspondence in Eq.~\ref{eq:app_interface_match_score} is computed against this expanded region rather than the raw $<$6\,\AA{} contact core.

The native region center is computed from the expanded support using Gaussian weights determined by the distance $d_j$ to the contact core,
\begin{equation}
    w_j \propto
    \exp\left[
        -\frac{1}{2}
        \left(
            \frac{d_j}{4\,\text{\AA}}
        \right)^2
    \right].
\end{equation}
The weighted C$\alpha$ centroid is used for the center-distance term in predicted/native interface matching.

\paragraph{Native-derived conditioning for baseline methods.}
The compared methods expose different interfaces for incorporating binding-site information.
We therefore preserve the conditioning format supported by each released inference pipeline rather than forcing all methods to consume an identical pocket representation.
In this context, native-derived interface conditioning refers to interaction information derived from the native complex and supplied in the form supported by the corresponding method.
For example, ODesign explicitly consumes a list of protein hotspot residues, whereas other methods use a receptor crop or do not take an explicit hotspot input.
The exact information available to each method is summarized below.

\subsection{Baseline Conditioning and Implementation Details}
\label{app:baseline_details}

To preserve the intended inference setting of each baseline, we follow the corresponding released generation pipeline instead of imposing a common artificial conditioning interface.
The target protein and requested RNA length are fixed consistently whenever supported by the method.
Table~\ref{tab:app_baseline_conditioning} summarizes the information used by each approach in the main comparison.

\begin{table}[!h]
    \centering
    \small
    \setlength{\tabcolsep}{4.5pt}
    \begin{tabular}{lcccc}
        \hline
        Method
        & Protein input
        & Site information
        & RNA information
        & Candidate selection \\
        \hline
        ODesign
        & Structure
        & Native-derived hotspot
        & Requested length
        & Single output \\

        RFDpoly
        & Structure
        & No explicit hotspot
        & Requested length
        & Single output \\

        RNAFlow
        & Structure crop
        & Native-RNA-guided crop
        & Requested length
        & Single output \\

        NACraft
        & Sequence
        & No explicit hotspot
        & Requested length
        & Highest model confidence \\

        RiboBridge
        & Structure
        & Predicted interface
        & Requested length
        & Matched hypothesis \\
        \hline
    \end{tabular}
    \caption{
    Conditioning information used by each method in the main comparison.
    For RiboBridge, ``Matched hypothesis'' means that predicted-interface candidates are generated independently before a correspondence with the native interface is established for evaluation.
    }
    \label{tab:app_baseline_conditioning}
\end{table}

\subsection{Structural Metrics and Failure Handling}
\label{app:structural_metrics}

\paragraph{Interface recovery and RMSD metrics.}
We evaluate protein--RNA complex recovery using the fraction of native contacts (fNAT), interface RMSD (iRMSD), ligand RMSD (lRMSD), and DockQ.
Let $\mathcal C_{\mathrm{nat}}$ denote the native interface-contact set and $\mathcal C_{\mathrm{pred}}$ the corresponding contact set in the generated complex.
The fraction of native contacts is
\begin{equation}
    \mathrm{fNAT}
    =
    \frac{
        |\mathcal C_{\mathrm{nat}}
        \cap
        \mathcal C_{\mathrm{pred}}|
    }{
        |\mathcal C_{\mathrm{nat}}|
    }.
    \label{eq:app_fnat}
\end{equation}
iRMSD measures structural deviation over the interface after interface superposition, whereas lRMSD measures the global displacement of the RNA ligand after receptor alignment.
The former emphasizes local interface geometry, while the latter is also sensitive to protein-relative rigid-body placement.

Following DockQ ~\citep{mirabello2024dockq}, these quantities are combined as
\begin{equation}
    \mathrm{DockQ}
    =
    \frac{1}{3}
    \left[
        \mathrm{fNAT}
        +
        \frac{1}{
            1+
            \left(
                \frac{\mathrm{iRMSD}}{1.5}
            \right)^2
        }
        +
        \frac{1}{
            1+
            \left(
                \frac{\mathrm{lRMSD}}{8.5}
            \right)^2
        }
    \right].
    \label{eq:app_dockq}
\end{equation}
As in the main paper, a generated complex with DockQ $\geq 0.23$ is counted as a successful recovery.
fNAT, iRMSD, lRMSD, DockQ, and the clash-related metric are macro-averaged over outputs for which the corresponding metric is defined.
DockQ success is instead reported over all 124 test targets, with missing or non-evaluable outputs counted as unsuccessful.

\paragraph{Atomic clashes.}
We additionally report a clash score to characterize the steric quality of generated complexes.
For two scored heavy atoms $i$ and $j$, the van der Waals overlap is defined as
\begin{equation}
    o_{ij}
    =
    r_i^{\mathrm{vdW}}
    +
    r_j^{\mathrm{vdW}}
    -
    \|\mathbf x_i-\mathbf x_j\|_2 .
\end{equation}
An atom pair is counted as a clash when
$o_{ij}>0.4\,\text{\AA}$.
The resulting score is normalized by the number of scored heavy atoms,
\begin{equation}
    \mathrm{ClashScore}
    =
    1000
    \frac{
        N_{\mathrm{clash}}
    }{
        N_{\mathrm{atom}}
    }.
\end{equation}
Hydrogen atoms are excluded.
Local covalent neighbors within the same or sequentially adjacent residues are not counted as steric clashes.

\paragraph{Canonical conformation metrics.}
For canonical-generator ablations, Atom11 RMSD measures recovery of the complete RNA atom11 geometry, while canonical C4$'$ RMSD characterizes backbone-level conformational recovery.
Because canonical RNA structures are defined only up to an arbitrary global rigid transform, the generated structure is first aligned to the native RNA using corresponding C4$'$ atoms through a proper Kabsch transformation.
RMSD is then computed after this alignment.
This alignment is used only for evaluating intrinsic RNA conformation and is not provided to the downstream interaction model.

\paragraph{Interaction-map metrics.}
For protein-anchor/RNA-C4$'$ pairs, F1@5\,\AA{} and F1@8\,\AA{} are computed using contact thresholds of 5\,\AA{} and 8\,\AA{}, respectively.
Distogram MAE converts each predicted finite-distance class into the corresponding bin midpoint and measures its deviation from the target distance,
\begin{equation}
    \mathrm{MAE}_{\mathrm{dist}}
    =
    \frac{1}{|\Omega_f|}
    \sum_{(j,i)\in\Omega_f}
    \left|
        \widehat d_{ji}
        -
        d^\star_{ji}
    \right|,
    \label{eq:app_distogram_mae}
\end{equation}
where $\Omega_f$ contains only pairs assigned to a finite target-distance class.
Pairs in the long-range class are excluded from this MAE.

\paragraph{Pose error and observability.}
To isolate rigid-placement error, let $\mathbf X_{\mathrm{oracle}}$ denote a generated canonical RNA transformed by its oracle rigid pose, and let $\mathbf X_{\mathrm{RBL}}$ denote the same canonical RNA transformed by the pose recovered by RBL.
Pose RMSD is computed directly between the corresponding C4$'$ coordinates in
$\mathbf X_{\mathrm{RBL}}$ and $\mathbf X_{\mathrm{oracle}}$ without any additional superposition.

A recovered pose is considered locally observable only when the finite-range constraints yield a full-rank six-dimensional pose Jacobian at the final solution.
If a pose cannot be recovered or is not locally observable, its DockQ is set to zero in the failure-aware stage-wise evaluation.
Metrics such as pose RMSD, which are defined only for an observable pose, are averaged only over the corresponding evaluable targets.

\subsection{Statistical Analysis and Bootstrap Confidence Intervals}
\label{app:statistical_analysis}

All confidence intervals are computed by resampling test targets while preserving the pairing of the same target across compared variants or adjacent structural stages.
Unless otherwise stated, we use $10{,}000$ bootstrap replicates ~\citep{efron1992bootstrap} and report the $2.5$th and $97.5$th percentiles of the bootstrap distribution as a two-sided $95\%$ confidence interval.

\paragraph{Canonical-generator ablations.}
The pocket-conditioned canonical generator is evaluated using two independently trained seeds.
For each target and training seed, we first compute the paired Full-versus-variant effect.
The two seed-specific effects are then averaged at the target level before bootstrap resampling.
The resulting confidence interval therefore quantifies uncertainty across test targets rather than treating the two training seeds as independent statistical samples.

To make the direction of the reported effects consistent across metrics, for a lower-is-better metric $m$ we define
\begin{equation}
    \Delta_m
    =
    m_{\mathrm{variant}}
    -
    m_{\mathrm{Full}},
\end{equation}
whereas for a higher-is-better metric we define
\begin{equation}
    \Delta_m
    =
    m_{\mathrm{Full}}
    -
    m_{\mathrm{variant}}.
\end{equation}
Thus, $\Delta_m>0$ consistently indicates an advantage of the Full model.
The confidence intervals reported in Table~\ref{tab:ablation_generator} follow this convention.

\paragraph{Stage-wise error decomposition.}
For two consecutive structural stages $s$ and $s+1$, the failure-aware effect for target $n$ is
\begin{equation}
    \Delta^{\mathrm{FA}}_n
    =
    Q_n^{(s)}
    -
    Q_n^{(s+1)},
\end{equation}
where $Q$ denotes DockQ.
If the RBL solution at a stage is not observable, the corresponding failure-aware DockQ is set to zero.
Target-paired bootstrap resampling of
$\{\Delta^{\mathrm{FA}}_n\}$
produces the $95\%$ confidence intervals shown for the filled markers in
Figure~\ref{fig:ablation_individual}.

We additionally report a jointly observable effect computed only over targets for which both neighboring stages yield observable poses,
\begin{equation}
    \Delta^{\mathrm{JO}}
    =
    \frac{1}{|\mathcal O_s\cap\mathcal O_{s+1}|}
    \sum_{n\in\mathcal O_s\cap\mathcal O_{s+1}}
    \left(
        Q_n^{(s)}-Q_n^{(s+1)}
    \right).
\end{equation}
This quantity separates structural degradation on jointly evaluable targets from losses caused by a change in pose observability.

\paragraph{Interaction-predictor ablations.}
Ablations of the cross-frame interaction model are evaluated on the same formally matched target cohort using paired comparisons.
Target-level metrics such as DockQ preserve target pairing.
Contact F1 is recomputed from the TP, FP, and FN counts accumulated over each bootstrap sample rather than averaging per-target F1 values.
Distogram MAE is aggregated over valid protein--RNA pairs.
This avoids an additional target-level averaging step when targets contain different numbers of pairwise interaction constraints.

\subsection{Sequence Design and AlphaFold~3 Self-consistency Protocol}
\label{app:af3_protocol}

Native-complex recovery alone does not establish that a generated RNA backbone can be realized by an RNA sequence.
We therefore evaluate sequence realizability and structural self-consistency through inverse folding followed by AlphaFold~3 refolding.
This evaluation is performed independently of the structural generation pipeline.
AlphaFold~3 is not given the generated three-dimensional RNA coordinates as input.

\paragraph{RNA sequence source.}
For ODesign, RNAFlow, and NACraft, which directly generate RNA sequences, we retain the sequence produced by the corresponding model.
RFDpoly and RiboBridge generate RNA backbones and therefore use the same NA-MPNN sequence-design protocol.
The sequence source used for each method is summarized in Table~\ref{tab:app_sequence_source}.

\begin{table}[!h]
    \centering
    \small
    \begin{tabular}{lc}
        \hline
        Method & RNA sequence source \\
        \hline
        ODesign    & Direct model output \\
        RFDpoly    & NA-MPNN \\
        RNAFlow    & Direct model output \\
        NACraft    & Direct model output \\
        RiboBridge & NA-MPNN \\
        \hline
    \end{tabular}
    \caption{
    RNA sequence sources used in the AlphaFold~3 self-consistency evaluation.
    }
    \label{tab:app_sequence_source}
\end{table}

\paragraph{NA-MPNN sequence design.}
For RFDpoly and RiboBridge, we use the same NA-MPNN design model.
The protein sequence is kept fixed and all RNA positions are designed.
Inference uses temperature $0.1$ and random seed 42.
One RNA sequence is generated for each backbone, without multi-sequence sampling or native-sequence-based reranking.
The native RNA sequence is therefore not used as a selection criterion during inverse folding.

\paragraph{AlphaFold~3 inputs.}
For each design, AlphaFold~3 receives the target protein sequence and designed RNA sequence.
The input does not contain the generated RNA backbone coordinates, native RNA coordinates, native complex pose, or geometric restraints derived from the main structural evaluation.
Protein MSA and template information are constructed independently by the standard AlphaFold~3 data pipeline, with a maximum template date of September 30, 2021.

Inference uses model seed 42, 10 recycles, and 5 diffusion samples.
For each target, we evaluate the top-ranked prediction produced by AlphaFold~3 itself.
No prediction is selected using native RMSD, self-consistency RMSD, or another ground-truth structural criterion.

\paragraph{Sequence recovery.}
RNA sequence recovery is computed from a global alignment between the designed sequence and the native RNA sequence.
Let $N_{\mathrm{match}}$ denote the number of identical nucleotides in the alignment and let $L_1$ and $L_2$ denote the two sequence lengths.
We define
\begin{equation}
    \mathrm{Rec.}
    =
    \frac{
        N_{\mathrm{match}}
    }{
        \max(L_1,L_2)
    }.
\end{equation}
The main paper reports the mean recovery over designs for which the sequence comparison is defined.

\paragraph{Conformational self-consistency.}
scRMSD evaluates whether the designed sequence refolds into the intended intrinsic RNA conformation.
Given corresponding C4$'$ atoms from the source RNA backbone and the AlphaFold~3 prediction, we first perform an RNA-only Kabsch alignment and then compute
\begin{equation}
    \mathrm{scRMSD}
    =
    \sqrt{
        \frac{1}{L}
        \sum_{i=1}^{L}
        \|
            \widehat{\mathbf x}^{\,\mathrm{AF3}}_i
            -
            \mathbf x^{\,\mathrm{design}}_i
        \|_2^2
    }.
\end{equation}
Because this metric removes the overall rigid-body difference between the two RNA structures, it primarily measures whether the designed sequence realizes the intended intrinsic conformation.

\paragraph{Complex self-consistency.}
scLRMSD evaluates whether the refolded RNA also preserves its designed pose relative to the target protein.
We first align the AlphaFold~3 protein to the design protein using corresponding C$\alpha$ atoms and apply the same rigid transformation to the AlphaFold~3 RNA.
RNA C4$'$ RMSD is then computed without any additional RNA alignment.
scLRMSD therefore reflects both changes in RNA conformation and changes in protein-relative placement.

For RNAFlow, whose released pipeline may operate on a local crop of the target protein, the cropped protein sequence is first mapped back to the corresponding region of the full AlphaFold~3 target protein before constructing the C$\alpha$ correspondence.

\paragraph{scTM and designability.}
We additionally report an RNA-only self-consistency TM score.
After RNA C4$'$ Kabsch alignment, the score is
\begin{equation}
    \mathrm{scTM}
    =
    \frac{1}{L}
    \sum_{i=1}^{L}
    \frac{
        1
    }{
        1+
        \left(
            d_i/d_0(L)
        \right)^2
    },
\end{equation}
where $d_i$ is the distance between corresponding C4$'$ atoms and $d_0(L)$ is a length-dependent normalization scale.
This metric complements scRMSD by measuring global fold-level self-consistency.

Following the definition used in the main paper, a design is considered designable when the AlphaFold~3 self-consistency evaluation succeeds and
\begin{equation}
    \mathrm{scRMSD}<2\,\text{\AA}.
\end{equation}
Designability is reported as the fraction of all 124 test targets satisfying this criterion.
It therefore reflects both failures in sequence design or refolding and the conformational fidelity of successfully refolded designs.

\section{Additional Analyses}
\label{app:additional_analyses}

This section provides additional analyses of the individual stages of RiboBridge.
We first examine whether the multi-pocket predictor preserves plausible native-interface
hypotheses when the binding site is unknown.
We then expand the stage-wise error decomposition and provide additional
ablations and diagnostics for canonical RNA generation, cross-frame interaction
prediction, and rigid placement.

\subsection{Interface Proposal Coverage and Matching Quality}
\label{app:pocket_coverage}

At inference time, RiboBridge does not know the native binding site.
The purpose of the pocket predictor is therefore not necessarily to collapse the full
protein surface into a single interface prediction, but to retain a compact set of
spatially distinct hypotheses that can be processed independently by downstream generation.

\paragraph{Proposal multiplicity.}
The pocket predictor starts from eight learned queries and applies residue-level filtering,
proposal-quality filtering, and non-maximum suppression before downstream generation.
Across the 124 test targets, the blind evaluation retains 202 interface hypotheses,
corresponding to an average of 1.63 proposals per target and a median of 2.
The number of retained proposals ranges from one to three.
Specifically, 60 targets retain one proposal, 50 retain two, and 14 retain three.

\begin{table}[!h]
    \centering
    \small
    \begin{tabular}{lc}
        \hline
        Quantity & Value \\
        \hline
        Raw learned queries / target & 8 \\
        Test targets & 124 \\
        Total retained proposals & 202 \\
        Mean proposals / target & 1.629 \\
        Median proposals / target & 2 \\
        Range & 1--3 \\
        Targets with 1 proposal & 60 \\
        Targets with 2 proposals & 50 \\
        Targets with 3 proposals & 14 \\
        \hline
    \end{tabular}
    \caption{
    Retained interface hypotheses in the blind 124-target evaluation.
    }
    \label{tab:app_pocket_count}
\end{table}

\paragraph{Formal interface coverage.}
We quantify whether the retained hypothesis set covers the native interaction region using
the formal matching criterion from \S\ref{app:candidate_matching}.
A predicted interface is formally matched when its Dice overlap with the expanded native
interface is at least 0.30 and the distance between the two interface centers is at most
6\,\AA{}.

Using only the highest-confidence proposal, 45 of 124 targets satisfy this criterion,
corresponding to Recall@1 of 36.29\%.
Allowing the top two retained proposals increases coverage to 53/124 (42.74\%),
and considering all retained hypotheses further increases coverage to 54/124 (43.55\%).

\begin{table}[!h]
    \centering
    \small
    \begin{tabular}{lcc}
        \hline
        Interface budget & Matched targets & Recall \\
        \hline
        Top-1 & 45/124 & 36.29\% \\
        Top-2 & 53/124 & 42.74\% \\
        All retained (Top-3) & 54/124 & 43.55\% \\
        \hline
    \end{tabular}
    \caption{
    Formal native-interface coverage as the retained proposal budget increases.
    }
    \label{tab:app_pocket_recall}
\end{table}

Among the 54 formally matched targets, the selected matching hypothesis is ranked first
for 45 targets, second for 8, and third for 1.
Thus, 9 of the 54 recoverable native-interface hypotheses would be missed if the predictor
committed to its top-ranked proposal alone.
In this evaluation, every matched target has exactly one retained proposal that
satisfies the formal criterion, and all stage-wise diagnostic assignments therefore use a
single predicted query.

\paragraph{Matching quality.}
The rank-1 predicted interface has a mean native-interface Dice of 0.503 and a median of
0.619.
Selecting the retained proposal with the largest native overlap increases these values to
0.573 and 0.647, respectively.
For the 54 formally matched interfaces, the mean Dice increases to 0.835 and the median to
0.862.
Their mean center distance is 2.73\,\AA{} and the median is 2.34\,\AA{}.

\begin{table}[!h]
    \centering
    \small
    \begin{tabular}{lccc}
        \hline
        Selection
        & Mean Dice
        & Median Dice
        & Median center dist. (\AA{}) \\
        \hline
        Rank-1 proposal
        & 0.503
        & 0.619
        & 10.72 \\

        Best-overlap retained proposal
        & 0.573
        & 0.647
        & 8.42 \\

        Formal matched proposal
        & 0.835
        & 0.862
        & 2.34 \\
        \hline
    \end{tabular}
    \caption{
    Correspondence between predicted interfaces and the expanded native
    interaction region.
    }
    \label{tab:app_pocket_matching_quality}
\end{table}

Proposal confidence is moderately associated with native-interface overlap
(Spearman $\rho=0.420$) and with the binary formal-match indicator
($\rho=0.465$).
By contrast, confidence rank has only a weaker association with native Dice
($\rho=-0.232$).
These results indicate that predictor confidence contains information about interface quality,
while leaving non-negligible uncertainty over which surface region corresponds to the native
binding site.
This motivates retaining several spatially distinct hypotheses rather than committing to a
single interface before RNA generation.

\paragraph{Independent pocket-predictor benchmark.}
The interface predictor achieves a top-1 pocket Dice/F1 of 0.833,
PR-AUC of 0.906, and ROC-AUC of 0.959 on its validation benchmark.
When selected proposals are considered jointly, the union Dice/F1 is 0.852 and the union
IoU is 0.742.
The pocket-count MAE is 0.231.
These residue-level and proposal-set measurements complement the blind test-time coverage
analysis above.

\paragraph{Summary.}
These results reveal two complementary properties of the interface predictor.
First, retaining multiple hypotheses improves native-interface coverage over committing to
the highest-confidence proposal alone, confirming that binding-site uncertainty remains
multimodal at inference time.
Second, when a predicted hypothesis satisfies the formal matching criterion, its interface
geometry is typically close to the native region.
Thus, the primary remaining challenge is not the geometric quality of a matched proposal,
but reliably covering and ranking the correct interface on the full protein surface.
This motivates the multi-hypothesis design of RiboBridge and the post-hoc
interface-correspondence protocol used to isolate downstream RNA generation
and assembly quality from interface ranking.

\subsection{Detailed Stage-wise Error Decomposition}
\label{app:error_decomposition_details}

The stage-wise analysis in Figure~\ref{fig:ablation_individual} progressively replaces
oracle information with the corresponding predicted component.
All stages use the same 54 targets satisfying the formal interface-matching criterion and
keep the selected predicted-pocket condition fixed.
The resulting sequence of interventions separates errors introduced by RNA representation,
canonical generation, interaction prediction, and rigid placement.

\paragraph{Stage definitions.}
Table~\ref{tab:app_error_stages} summarizes the information used at each stage.
The native/native sanity check evaluates the native complex itself.
GT-token reconstruction first encodes the native RNA into the RiboSphere structural-token
space and reconstructs it with the frozen decoder before applying an oracle C4$'$ Kabsch
placement.
The generated-canonical stage replaces these reconstructed tokens with the output of the
pocket-conditioned canonical generator while retaining the same oracle placement.

The remaining stages keep this generated RNA conformation fixed.
The exact-distance stage computes protein--RNA cross-frame distances at the oracle pose and
asks the reference RBL solver to recover that pose.
The oracle-37-bin stage discretizes the same exact ranges using the interaction predictor distance
classes before placement.
The predicted-map stage replaces these oracle constraints with the output of the learned
interaction predictor.
The final stage differs only in enabling the nonlinear fallback when the reference
linear initialization is rank deficient.

\begin{table*}[!h]
    \centering
    \small
    \setlength{\tabcolsep}{5pt}
    \begin{tabular}{lllll}
        \hline
        Stage
        & RNA source
        & Interaction
        & Placement
        & Oracle input \\
        \hline

        Native/native
        & Native
        & Native
        & Native
        & Full complex \\

        GT-token recon.
        & GT-token recon.
        & --
        & Oracle pose
        & Tokens + pose \\

        Generated canonical
        & Generated
        & --
        & Oracle pose
        & Pose \\

        Exact dist. + RBL
        & Generated
        & Exact distances
        & RBL
        & Distances \\

        Oracle 37-bin + RBL
        & Generated
        & Oracle bins
        & RBL
        & Distance bins \\

        Predicted map
        & Generated
        & Predicted
        & RBL
        & None \\

        Predicted map + fallback
        & Generated
        & Predicted
        & RBL + fallback
        & None \\
        \hline
    \end{tabular}
    \caption{
    Stages used for error decomposition.
    Consecutive stages replace one source of oracle information with its predicted counterpart.
    }
    \label{tab:app_error_stages}
\end{table*}

\paragraph{Absolute stage-wise behavior.}
GT-token reconstruction preserves substantially more of the native RNA geometry than direct
canonical generation.
Its canonical C4$'$ RMSD is 1.32\,\AA{} and its failure-aware DockQ is 0.854.
Replacing reconstructed native tokens with generated canonical RNA increases C4$'$ RMSD to
7.73\,\AA{} and reduces DockQ to 0.464.
This transition is the largest single loss in the decomposition.

When exact cross-frame distances are supplied, reference RBL remains observable for 53 of
54 targets and recovers the oracle pose with a mean pose RMSD of
0.0108\,\AA{} and a median of 0.0065\,\AA{}.
Discretizing the same distances into the 37-bin representation leaves observability unchanged,
with a median pose RMSD of only 0.100\,\AA{}.

The major downstream degradation occurs when the oracle interaction constraints are replaced by predicted interaction constraints.
Reference-RBL observability decreases from 53/54 to 44/54, and the median pose RMSD among
observable targets increases to 21.02\,\AA{}.
Failure-aware DockQ decreases from 0.445 to 0.253.
Nonlinear fallback recovers three additional observable targets, increasing observability from
81.5\% to 87.0\%, but changes aggregate DockQ only marginally.

\begin{table*}[!h]
    \centering
    \small
    \setlength{\tabcolsep}{4pt}
    \begin{tabular}{lcccccc}
        \hline
        Stage
        & Obs.
        & DockQ
        & fNAT
        & iRMSD
        & lRMSD
        & Pose RMSD (mean/med.) \\
        \hline
        Native/native
        & 54/54
        & 1.000
        & 1.000
        & $\approx0$
        & 0
        & 0/0 \\

        GT-token recon.
        & 54/54
        & 0.854
        & 0.834
        & 0.858
        & 1.319
        & 0/0 \\

        Generated RNA
        & 54/54
        & 0.464
        & 0.461
        & 4.183
        & 7.728
        & 0/0 \\

        Exact dist. + RBL
        & 53/54
        & 0.446
        & 0.441
        & 4.261
        & 7.804
        & 0.011/0.007 \\

        Oracle bins + RBL
        & 53/54
        & 0.445
        & 0.453
        & 4.261
        & 8.217
        & 0.888/0.100 \\

        Predicted map
        & 44/54
        & 0.253
        & 0.232
        & 4.206
        & 22.869
        & 21.920/21.021 \\

        Pred. map + fallback
        & 47/54
        & 0.254
        & 0.232
        & 4.369
        & 24.599
        & 23.640/23.496 \\
        \hline
    \end{tabular}
    \caption{
    Stage-wise diagnostics on the 54-target formally matched subset.
    DockQ is failure-aware; all RMSDs are in \AA{}.
    }
    \label{tab:app_error_absolute}
\end{table*}

\paragraph{Paired transition effects.}
The transition from GT-token reconstruction to generated canonical RNA produces the largest
paired degradation, with a mean DockQ loss of 0.389
(95\% CI $[0.329,0.449]$).
Replacing the oracle 37-bin interaction geometry with predicted interaction geometry produces the second largest loss, 0.192
(95\% CI $[0.150,0.234]$).

By contrast, replacing exact distances with oracle 37-bin distances changes DockQ by only
0.0016, and its confidence interval includes zero.
For targets that remain observable before and after exact-distance RBL, the mean DockQ loss
introduced by the solver is only 0.00057.
These paired interventions identify canonical generation and interaction prediction as the
dominant error sources in the current pipeline.

\begin{table*}[!h]
    \centering
    \small
    \begin{tabular}{lccc}
        \hline
        Transition
        & Isolated source
        & $\Delta$DockQ loss (95\% CI)
        & $\Delta$ observable \\
        \hline
        Native $\rightarrow$ GT-token recon.
        & RiboSphere reconstruction
        & 0.1465 $[0.1193,0.1764]$
        & 0 \\

        GT-token $\rightarrow$ generated canonical
        & Canonical generation
        & 0.3893 $[0.3290,0.4487]$
        & 0 \\

        Oracle pose $\rightarrow$ exact-distance RBL
        & Rigid placement
        & 0.0181 $[0.0000,0.0537]$
        & $-1$ \\

        Exact distance $\rightarrow$ oracle 37-bin
        & Distance discretization
        & 0.0016 $[-0.0056,0.0101]$
        & 0 \\

        Oracle 37-bin $\rightarrow$ predicted map
        & Interaction prediction
        & 0.1920 $[0.1502,0.2335]$
        & $-9$ \\

        Reference RBL $\rightarrow$ fallback
        & Nonlinear recovery
        & $-0.0018$ $[-0.0043,0.0000]$
        & $+3$ \\
        \hline
    \end{tabular}
    \caption{
    Target-paired changes between consecutive stages.
    Positive $\Delta$DockQ denotes degradation from the upstream to downstream stage.
    }
    \label{tab:app_error_transitions}
\end{table*}

\paragraph{Per-target relationships.}
Canonical C4$'$ RMSD is strongly negatively associated with oracle-pose DockQ
(Spearman $\rho=-0.923$).
After interaction prediction and placement, canonical RMSD remains associated with final
DockQ ($\rho=-0.798$).
Among observable targets, pose RMSD is also strongly negatively associated with final DockQ
($\rho=-0.875$).
Interaction-map F1@8 correlates positively with final DockQ
($\rho=0.646$), while distogram MAE correlates with pose RMSD
($\rho=0.848$ among observable targets).
These relationships are consistent with the stage-wise decomposition, while remaining
descriptive rather than causal.

\paragraph{Summary.}
Together, these interventions localize the dominant errors in RiboBridge to canonical RNA
generation and cross-frame interaction prediction.
By contrast, distance discretization introduces little systematic degradation, and RBL
nearly recovers the oracle pose when accurate cross-frame constraints are available.
These results support the proposed decomposition: RNA conformation and protein-relative
assembly can be modeled separately, while the remaining challenge lies primarily in predicting
high-quality RNA structures and informative cross-frame interaction geometry.

\subsection{Qualitative Complex Recovery and Error Modes}
\label{app:qualitative_recovery}

\begin{figure*}[!t]
    \centering
    \includegraphics[width=\textwidth]{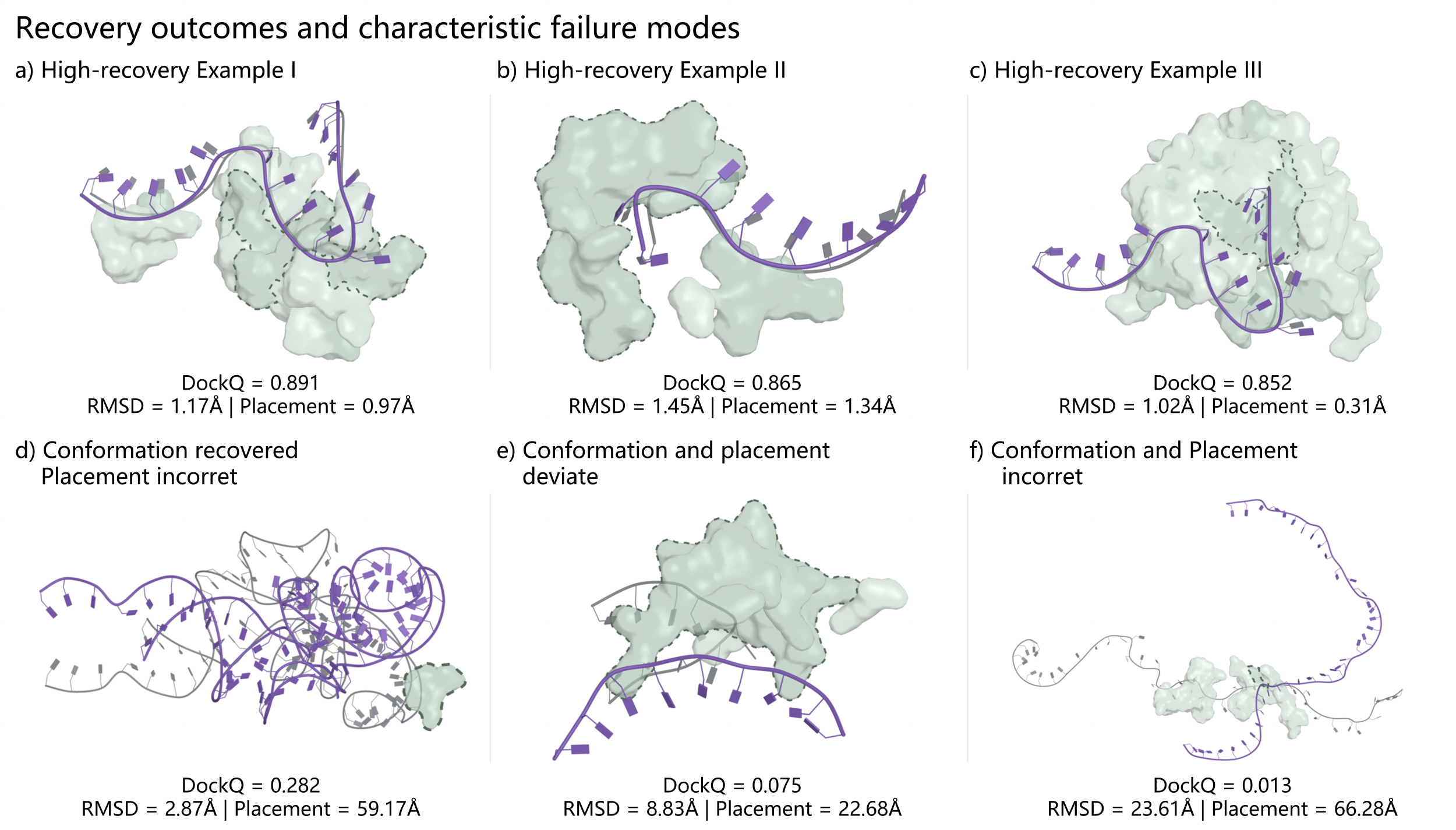}
    \caption{
    \textbf{Qualitative examples of complex recovery and
    conformation--placement errors.}
    \textbf{(a--c)} High-recovery complexes.
    \textbf{(d--f)} Examples with different degrees of RNA
    conformational and protein-relative pose deviation.
    Generated RNA is purple, native RNA is gray, and the local
    protein environment is pale green.
    RMSD measures RNA conformational deviation after alignment;
    placement error measures deviation of the recovered pose
    without an additional RNA superposition.
    DockQ evaluates recovery of the native complex.
    }
    \label{fig:app_qualitative_recovery}
\end{figure*}

Figure~\ref{fig:app_qualitative_recovery} complements the stage-wise
analysis with examples of final protein--RNA complexes. Panels (a--c)
show high-recovery cases, in which both RNA conformation and
protein-relative placement are close to their reference configurations.
Their DockQ values range from 0.852 to 0.891.

The remaining panels illustrate how these two deviations can coexist
to different degrees. In panel (d), RNA conformation remains
comparatively close to native, whereas its protein-relative pose
differs substantially; the resulting DockQ is 0.282. Panel (e)
shows appreciable deviations in both conformation and placement,
and panel (f) shows large deviations in both, with DockQ values of
0.075 and 0.013, respectively. These examples visualize the outcomes
summarized by the quantitative error decomposition. A pose discrepancy
in a final complex reflects the combined effects of predicted
cross-frame constraints and rigid placement; it does not, by itself,
identify an error in the rigid-placement solver.

\subsection{Canonical RNA Generation Ablation Details}
\label{app:generator_ablation_details}

The canonical-generator ablation compares four independently trained variants:
the Full model with matched predicted-pocket conditioning, a model without self-correction, a length-only model,
and a shuffled-pocket control.
All variants share the same architecture family, training split, optimization budget,
checkpoint-selection procedure, and evaluation cohort unless explicitly stated otherwise.

\paragraph{Variant construction.}
The Full model uses the formally matched predicted-pocket condition together with the
self-correction procedure described in Appendix~A.
For the No self-correction variant, the on-policy rollout branch is disabled during training.
This variant is trained independently and is not obtained by simply disabling a loss term at
inference time.

For Length-only, the protein/pocket condition is replaced by a single valid
zero-valued 512-dimensional conditioning token.
The requested RNA length remains available through sequence length and valid-position masks,
but no target-specific protein or interface representation is supplied.

For Shuffled pocket, each training example retains its original RNA structural target and
requested length but receives a pocket condition from another target with exactly the same RNA
length.
The donor assignment is fixed per training seed, excludes self-donation, prefers a different
PDB entry when possible, and minimizes pocket-size mismatch within each exact-length group.
The control therefore preserves the presence and approximate size of the conditioning sequence
while breaking its correspondence with the target RNA conformation.

\paragraph{Evaluation protocol.}
All four variants are evaluated on the same 54 formally matched targets with identical frozen
pocket assignments, requested RNA lengths, generation settings, and evaluation seed.
Generated canonical structures are placed by an oracle proper Kabsch transform before
interface and complex metrics are computed.
The interaction predictor and RBL are not invoked.
The resulting iRMSD, fNAT, and DockQ therefore primarily measure differences in canonical RNA
conformation rather than downstream interaction prediction or rigid placement.

\paragraph{Absolute and paired results.}
Averaged over seeds 42 and 111, the Full model obtains an Atom11 RMSD of
7.972\,\AA{} and DockQ of 0.4691.
Removing self-correction changes the point estimates only slightly.
In contrast, removing target-specific interface conditioning substantially increases structural
error.
Length-only and Shuffled-pocket yield Atom11 RMSDs of
9.459\,\AA{} and 9.420\,\AA{} and reduce DockQ to
0.3622 and 0.3745, respectively.

The target-paired bootstrap intervals show the same pattern.
The Full-versus-Length-only DockQ advantage is 0.1069
with 95\% CI $[0.051,0.167]$, and the Full-versus-Shuffled advantage is
0.0946 with 95\% CI $[0.049,0.144]$.
In contrast, the Full-versus-No-self-correction difference is only 0.0032 with CI
$[-0.019,0.026]$.
The corresponding intervals for the structural RMSD and interface-recovery metrics follow the
same qualitative trend.

\paragraph{Summary.}
These results show that canonical RNA generation is not explained by RNA length alone:
target-specific interface information provides a substantial and statistically supported
signal for shaping the generated RNA conformation.
The shuffled-pocket control further indicates that the gain arises from the correspondence
between the interface and the RNA target rather than from additional conditioning tokens
alone.
This supports the use of pocket-conditioned canonical generation before cross-frame assembly.
By contrast, the current experiments do not establish a stable aggregate benefit from
self-correction.

\subsection{Cross-frame Interaction Ablation Details}
\label{app:interaction_ablation_details}

We ablate the geometric components of the cross-frame interaction predictor
while keeping the canonical RNA inputs and the reference rigid-placement
solver without nonlinear fallback fixed.

\paragraph{Local geometry.}
The Full model uses explicit node-level shape descriptors and frame-aware within-chain edge
features.
The node descriptor contains the mean and standard deviation of eight-nearest-neighbor
distances, two covariance-eigenvalue ratios, local bend, and a frame-validity indicator.
Within-chain edges include normalized displacement directions represented in the local frames
of both endpoints together with frame-validity indicators.

The \emph{without local geometry} variant zeros these explicit node-shape and edge-orientation
features.
It retains scalar sequence/radial information, invariant pairwise distances, and the geometry
propagated by the remaining message-passing layers.
This ablation therefore removes explicit local-frame geometry rather than all geometric
information.

\paragraph{Triangle reasoning.}
The Full model processes the protein--RNA pair grid using 20 triangle reasoning stages.
Each stage first applies local $1\times9$, $3\times3$, and $9\times1$ residual operations,
followed by triangle aggregation using within-protein and within-RNA geometric relations.
In the \emph{without triangle reasoning} variant, the Inception operations are retained while
the triangle aggregation module is replaced by a parameter-matched pointwise pair module.
This isolates shared-axis higher-order reasoning while keeping pair-grid capacity broadly
comparable.

\paragraph{Contact supervision.}
The Full model is trained using the 37-class distogram cross-entropy together with balanced
$<$5\,\AA{} and $<$8\,\AA{} contact objectives.
The CE-only variant retains the complete architecture but removes the explicit contact-loss
terms, leaving only the multiclass distogram objective.

All variants use the same frozen canonical RNA inputs and the same training and validation
targets.

\paragraph{Ablation behavior.}
As reported in the main paper, the Full model achieves F1@5\,\AA{} of 0.272,
F1@8\,\AA{} of 0.388, and distogram MAE of 3.571\,\AA{}.
Removing explicit local geometry reduces F1@5\,\AA{} and F1@8\,\AA{} to
0.235 and 0.331 and increases distogram MAE to 3.793\,\AA{}.
Removing triangle reasoning has a smaller effect at 5\,\AA{} but reduces
F1@8\,\AA{} to 0.348.
CE-only yields a distogram MAE close to that of the Full model
(3.573 versus 3.571\,\AA{}) while reducing F1@5\,\AA{} to 0.227.

The similar average distance error of Full and CE-only does not imply equivalent
short-range interaction quality.
In the contact predictions, the Full model produces substantially fewer false-positive
contacts.
At the 5\,\AA{} threshold, false positives decrease from 47 for CE-only to 18 for
Full; at 8\,\AA{}, they decrease from 213 to 94.
Thus, explicit contact supervision primarily improves the short-range part of the interaction
map that is most relevant for constraining complex placement, even when the average
finite-distance error changes little.

Target-paired analysis similarly shows clearer effects on interaction-map quality than on final
placement quality.
The Full model improves F1@8 over the variants without local geometry or triangle reasoning,
and improves both F1@5 and F1@8 over CE-only.
By comparison, the downstream pose and DockQ differences are smaller and more variable across
targets.
This suggests that the ablated geometric components primarily improve the predicted
cross-frame constraints, while their conversion into an accurate rigid pose remains dependent
on the geometric informativeness of those constraints.

\subsection{Rigid-placement Observability and Fallback Analysis}
\label{app:rbl_analysis}

The stage-wise results indicate that the rigid-placement solver contributes little structural
error when sufficiently accurate cross-frame constraints are available.
We therefore analyze separately the conditions under which the localization problem is
observable and the role of the nonlinear fallback.

\paragraph{Linear solvability and local observability.}
RBL uses two distinct rank conditions.
The centered squared-range initialization estimates 12 linear unknowns and requires
\begin{equation}
    \operatorname{rank}(B)\geq12.
\end{equation}
Failure of this condition indicates that the reference linear initializer is underdetermined;
it does not by itself imply that no locally observable rigid pose exists.

After pose optimization, final local observability is evaluated using the six-dimensional
pose Jacobian $J$.
A recovered pose is accepted as locally observable only if
\begin{equation}
    \operatorname{rank}(J)=6.
\end{equation}
The nonlinear fallback can therefore recover a target that fails the linear
$\operatorname{rank}(B)\geq12$ condition as long as the resulting pose satisfies the final
six-degree-of-freedom observability test.

\paragraph{Behavior with exact interaction geometry.}
When exact cross-frame ranges are supplied, RBL returns an observable pose for 53 of the
54 diagnostic targets.
The mean and median pose RMSDs are only
0.0108\,\AA{} and 0.0065\,\AA{}.
On the 53 targets that remain jointly observable, the mean DockQ loss relative to the oracle
pose is only 0.00057.
This provides a direct sanity check that RBL nearly recovers the oracle rigid transform when
the interaction ranges are sufficiently informative.

The single exact-distance failure has 26 finite ranges but a linear-system rank of 9,
showing that accurate distances alone cannot determine all rigid-body degrees of freedom if
their geometric arrangement is insufficient.

\paragraph{Effect of distance discretization.}
Replacing exact ranges with their oracle 37-bin representation preserves the same
53/54 observable targets.
Failure-aware DockQ changes by only 0.0016, with a paired confidence interval spanning zero.
Although mean pose RMSD increases from 0.011\,\AA{} to
0.888\,\AA{}, the median remains 0.100\,\AA{}.
The larger mean is therefore driven by a small number of outliers rather than by a systematic
degradation across the diagnostic cohort.

\paragraph{Nonlinear fallback.}
With predicted interaction constraints, the reference solver is observable for 44 of 54
targets.
Ten targets fail the reference linear initialization and invoke nonlinear fallback.
The fallback recovers three additional targets, raising final observability to 47/54.

The recovered cases nevertheless remain structurally inaccurate, with pose RMSDs of
29.1\,\AA{}, 73.6\,\AA{}, and 43.9\,\AA{} and low DockQ values.
Accordingly, failure-aware DockQ changes only from 0.2525 to 0.2543 after fallback.
The fallback therefore improves geometric solvability but cannot compensate for inaccurate
interaction constraints.

\paragraph{Constraint availability and observability.}
Final observability is positively associated with the number of usable finite-range constraints
(Spearman $\rho=0.571$).
Observable targets have a median of 117 usable finite ranges, whereas the median for
unobservable targets is zero.
Associations with RNA length, predicted pocket size, contact F1, and distogram MAE are much
weaker in this cohort.
These observations support the narrower conclusion that insufficient finite-range support is
an important contributor to lost pose observability, without implying that constraint count
alone determines pose accuracy once the problem is observable.

\section{Computational Resources and Reproducibility}
\label{app:compute}

This section summarizes the computational resources, runtime and software environment associated with the reported RiboBridge experiments.
All experiments were conducted on a server equipped with eight NVIDIA RTX A6000 GPUs.
RiboSphere is used as a frozen pretrained representation and is therefore not included in the
training cost of RiboBridge itself.

\subsection{Training Cost}
\label{app:training_cost}

RiboBridge consists of three trainable components, while rigid placement is parameter-free.
Table~\ref{tab:app_training_cost} summarizes their model sizes and training costs.

\begin{table*}[!htbp]
    \centering
    \small
    \begin{tabular}{lcccc}
        \hline
        Component
        & Trainable params.
        & Hardware
        & Training budget
        & Training time \\
        \hline
        Interface predictor
        & 102.66M
        & 8 GPUs
        & 20 epochs
        & $\sim$4 h \\

        Canonical generator
        & 38.5M
        & 8 GPUs
        & 40 epochs
        & $\sim$4 h \\

        Interaction predictor
        & 13.3M
        & 8 GPUs
        & 50 epochs
        & $\sim$13 h \\
        \hline
    \end{tabular}
    \caption{
    Training cost of the three trainable components in RiboBridge.
    }
    \label{tab:app_training_cost}
\end{table*}

\subsection{Inference Cost}
\label{app:inference_cost}

Table~\ref{tab:app_inference_cost} summarizes the end-to-end inference cost measured on the blind 124-target evaluation, including pocket prediction and all retained interface candidates.

\begin{table}[!htbp]
    \centering
    \small
    \begin{tabular}{lc}
        \hline
        Quantity & Runtime \\
        \hline
        Full 124-target run & 1872.8 s \\
        Mean / target & 15.10 s \\
        Median / target & 11.01 s \\
        10th percentile & 6.06 s \\
        90th percentile & 28.33 s \\
        Median, 1 retained pocket & 6.47 s \\
        Median, 2 retained pockets & 11.55 s \\
        Median, 3 retained pockets & 30.44 s \\
        \hline
    \end{tabular}
    \caption{
    End-to-end inference cost on the blind 124-target evaluation.
    Runtime includes pocket prediction and downstream generation for all retained interface hypotheses.
    }
    \label{tab:app_inference_cost}
\end{table}

\end{document}